\documentclass[manuscript,screen]{acmart}
\AtBeginDocument{%
  }

\usepackage{color, xcolor} 
\usepackage{framed}
\usepackage{soul} 
\usepackage{graphicx}
\usepackage{subfigure}
\usepackage{arydshln}
\usepackage[most]{tcolorbox}
\usepackage{multirow}
\tcbuselibrary{breakable, skins} % 允许分页
\usepackage{colortbl}
\usepackage{algorithmicx,algorithm}
\usepackage[noend]{algpseudocode}
\definecolor{lightgray}{rgb}{0.88, 0.92, 0.98}
\definecolor{defblue}{rgb}{0.1843, 0.3333, 0.6}
\definecolor{defred}{rgb}{0.88, 0.2510, 0.3294}
\usepackage{booktabs}

\begin{document}

\title{Dual Knowledge-Enhanced Two-Stage Reasoner for Multimodal Dialog Systems}

\author{Xiaolin Chen}
\email{e-mail: cxlicd@gmail.com}
\affiliation{%
  \institution{National University of Singapore}
  \country{Singapore}
  }

\author{Xuemeng Song$^*$}
\affiliation{
  \institution{Southern University of Science and Technology}
  \country{China}
  }
\email{sxmustc@gmail.com}

\author{Haokun Wen}
\affiliation{%
  \institution{Harbin Institute of Technology (Shenzhen)}
  \country{China}
}
 \email{whenhaokun@gmail.com}

\author{Weili Guan}
\email{honeyguan@gmail.com}
\affiliation{
  \institution{Harbin Institute of Technology (Shenzhen)}
  \country{China}
}

\author{Xiangyu Zhao}
\email{xianzhao@cityu.edu.hk}
\affiliation{
  \institution{City University of Hong Kong}
  \country{China}
}

\author{Liqiang Nie$^*$}
\affiliation{%
  \institution{Harbin Institute of Technology (Shenzhen)}
  \country{China}
  }
 \email{nieliqiang@gmail.com}

\thanks{$^*$Corresponding authors: Xuemeng Song and Liqiang Nie.}

% \affiliation{%

% \affiliation{%

% \affiliation{%

% \affiliation{%

% \affiliation{%

% \affiliation{%

% \affiliation{%

% \affiliation{%

\begin{abstract}
Textual response generation is pivotal  for multimodal \mbox{task-oriented} dialog systems, which aims to generate proper  textual responses based on the multimodal context.  
While  existing efforts have demonstrated  remarkable  progress, there still exist the   following limitations: 1) \textit{neglect of unstructured review knowledge} and 2) \textit{underutilization of large language models (LLMs)}.
Inspired by this, we aim to fully utilize dual knowledge  (\textit{i.e., } structured attribute  and
unstructured review knowledge) with LLMs to promote
textual response generation  in multimodal task-oriented dialog systems.
 However, this task is non-trivial due to two key challenges:  1) \textit{dynamic knowledge type selection} and 2) \textit{intention-response decoupling}.
To address these challenges, we propose a novel dual knowledge-enhanced two-stage reasoner by adapting LLMs for  multimodal dialog systems (named  DK2R).
To be specific, DK2R first  extracts both structured attribute and unstructured review knowledge from external knowledge base given the dialog context. 
Thereafter,  DK2R uses an LLM to evaluate each knowledge type's utility by analyzing LLM-generated provisional probe responses.
Moreover,   DK2R  separately summarizes the intention-oriented key clues via dedicated reasoning, which are further  used as auxiliary signals to enhance LLM-based textual response generation.
Extensive experiments conducted on a public dataset verify  the superiority of DK2R. We have released the codes and parameters.
\end{abstract}



\maketitle

\section{Introduction}
\begin{figure}[!t]
    \centering
    \includegraphics[scale=0.45]{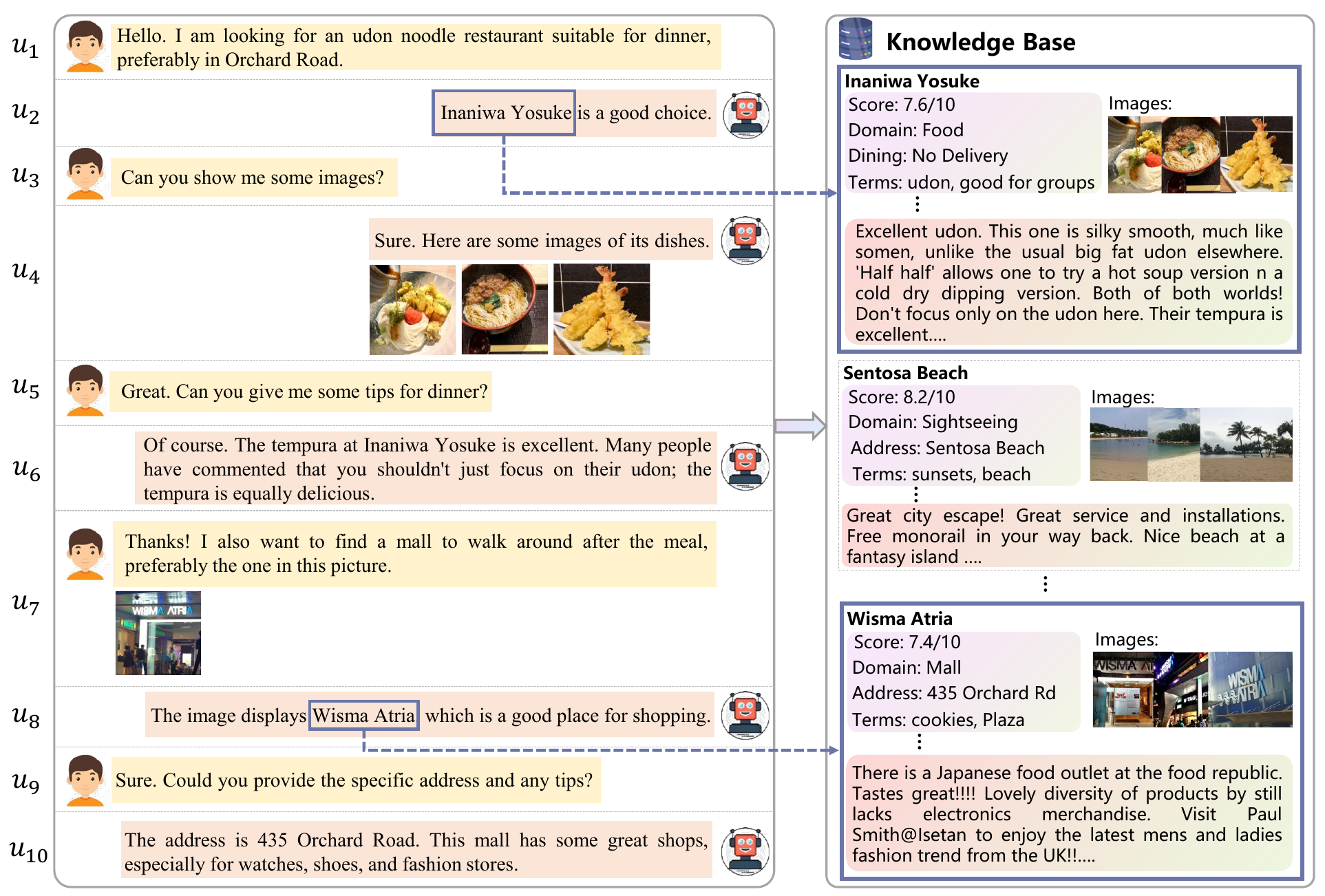}
    \caption{Illustration of a multimodal dialog system. ``$u$'': utterance. 
    } % The blue and green shaded  regions denote context-related structured attribute knowledge and unstructured review knowledge, respectively.
    \vspace{-1em}
    \label{figure_context}
\end{figure}

According to  statistics from Grand View Research\footnote{https://www.grandviewresearch.com/industry-analysis/conversational-ai-market-report.}, the global conversational AI market size was estimated at USD 11.58 billion in 2024 and is anticipated to grow rapidly.
Given the substantial economic value, task-oriented dialog systems, which are designed to  solve domain-specific tasks such as booking restaurant tables and making hotel reservations, have been attracting increasing attention.
Early studies predominantly  focus on \mbox{single-modal}, text-based interactions between users and agents.
However, recognizing the inherent inflexibility of such systems when applied to real-world scenarios, recent research efforts have  increasingly explored  multimodal task-oriented dialog systems.
As shown in Figure~\ref{figure_context}, these systems support multimodal interactions involving  both text and images. For example, to fulfill the user's request, the agent shows images of several dishes alongside a textual response in utterance $u_4$. Meanwhile, the user provides a picture to specify the desired shopping mall in utterance $u_7$.

 Multimodal task-oriented dialog systems involve two sub-tasks: textual response generation and image response selection.
In this work,  we focus on textual response generation, which is considered the more challenging task, following prior studies~\cite{10.1145/3343031.3350923,DBLP:conf/acl/MaL0C22,DBLP:conf/sigir/ChenSWNC23,DBLP:journals/tomccap/ChenSZWNC25}.
Although existing efforts have made significant progress, they have two key limitations. 1) \textbf{Neglect of unstructured review knowledge.} Current systems rely solely on the multimodal dialog context together with structured attribute knowledge (\textit{i.e., } attributes of entities) from an external knowledge base for response generation,  ignoring the vital role of unstructured review  knowledge (\textit{i.e., } the user experiential feedback of entities). Yet, such review knowledge is  also essential for generating informative responses. For instance, as exhibited in Figure~\ref{figure_context}, only by incorporating user reviews for ``Inaniwa Yosuke'',  can the system provide dining tips and generate the appropriate response as utterance $u_6$. 2) \textbf{Underutilization of LLMs.} Despite remarkable advances in pretrained large language models~(LLMs) like GPT-4~\cite{achiam2023gpt} and PaLM-2~\cite{anil2023palm}, existing multimodal task-oriented dialog systems largely depend on conventional generative models like BART~\cite{DBLP:conf/acl/LewisLGGMLSZ20} for textual response generation, leaving the potential of LLMs in this domain underexplored, particularly their powerful natural language processing and reasoning capabilities. 

To address these limitations, we propose to synergistically integrate both structured attribute and unstructured review knowledge with advanced LLMs for multimodal task-oriented dialog systems. 
However,  this integration presents two fundamental challenges.
{\textbf{1) Dynamic knowledge type selection.}} User intentions exhibit significant diversity, necessitating context-aware selection of relevant knowledge types.   
For example, as illustrated in Figure~\ref{figure_context}, generating response $u_{10}$ requires integrating both structured attribute knowledge (\textit{i.e., } the address attribute) and unstructured review knowledge, while response $u_6$ relies solely on review knowledge. 
Incorporating  irrelevant types of knowledge  (\textit{e.g., } using structured attribute knowledge when generating the response $u_6$) introduces noise that degrades generation quality.  
Nevertheless, existing datasets lack explicit annotations of knowledge-type utility, making the automatic identification of contextually relevant knowledge a critical challenge. 
{\textbf{2) Intention-response decoupling.}} 
A direct application of LLMs in our context would process the multimodal context and dual knowledge (both structured and unstructured) in a single end-to-end generation step via instruction tuning~\cite{DBLP:journals/corr/abs-2407-04973}. However, this approach conflates two distinct cognitive processes: (1) intention understanding and (2) knowledge-enhanced response generation. While external knowledge significantly enriches response generation, its  incorporation during intention reasoning may distort user intent interpretation. Thus, a core challenge is decoupling intention understanding from knowledge-enhanced response generation to ensure accurate user intention capture while effectively utilizing supplementary knowledge.

\begin{figure*}[!t]
    \centering
    \includegraphics[scale=0.36]{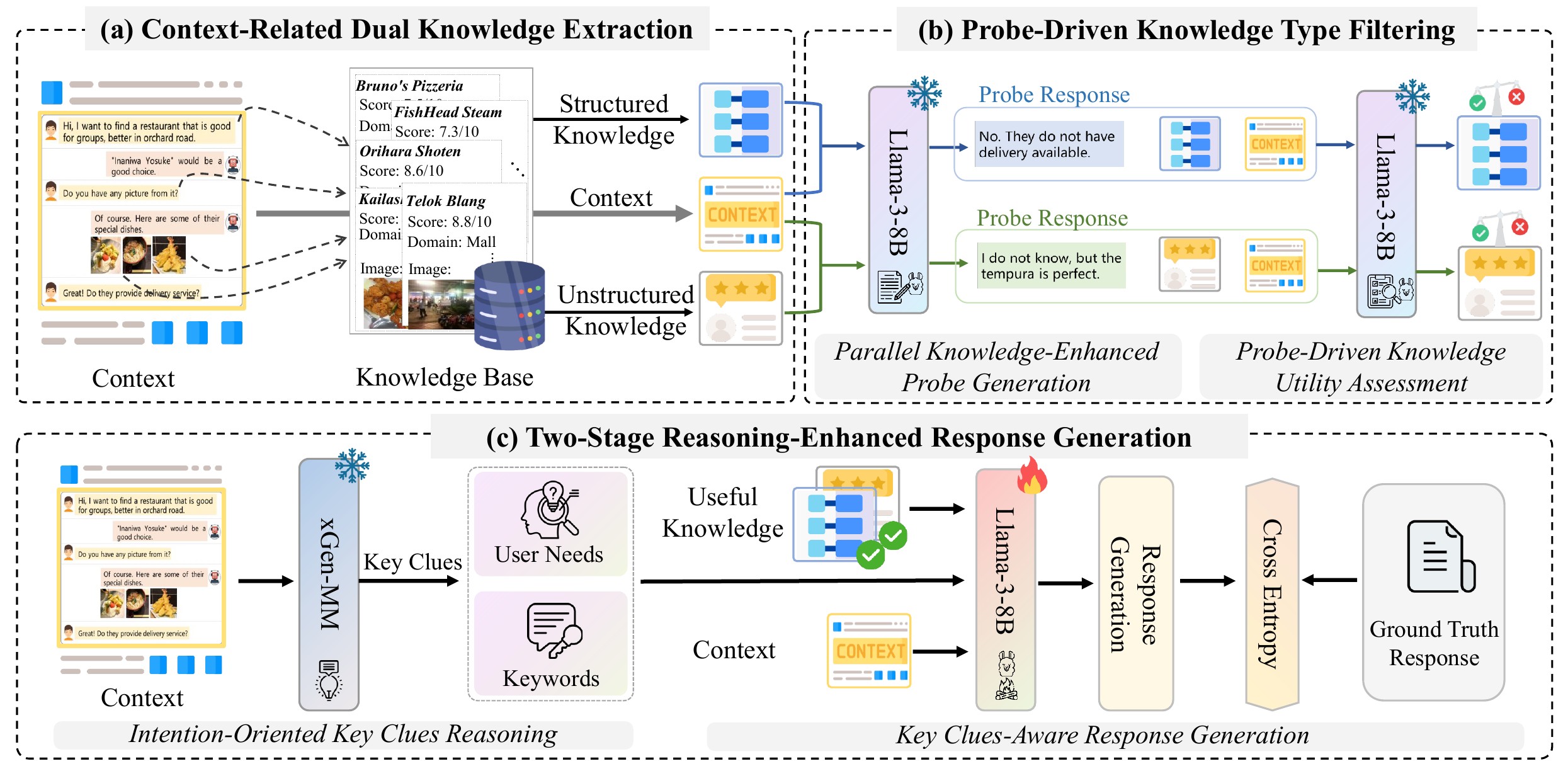}
        % \vspace{-0.2em}
    \caption{Illustration of the proposed model for multimodal task-oriented  dialog systems, which consists of  three key  components: \emph{context-related dual knowledge extraction}, \emph{probe-driven knowledge type filtering}, and \emph{two-stage reasoning-enhanced response generation}.}
    \label{figure_model}%\vspace{-0.8em}
\end{figure*}

To tackle  these challenges,  we  propose DK2R, a novel dual knowledge-enhanced two-stage reasoner for textual response generation in multimodal dialog systems.
As depicted in Figure~\ref{figure_model}, DK2R comprises  three key  components:  \textit{context-related dual knowledge extraction}, \textit{probe-driven knowledge type filtering}, and \textit{two-stage reasoning-enhanced response generation}.
Specifically, the context-related dual knowledge extraction component  retrieves relevant structured attribute and unstructured review knowledge conditioned on the given multimodal context.
Building upon this,  the probe-driven knowledge type filtering  component   first generates separate provisional probe responses for each knowledge type (\emph{i.e., }structured attribute and unstructured review) using the LLM, and then independently evaluates their utility through corresponding responses. 
Moreover, the two-stage reasoning-enhanced response generation component  decouples intention reasoning from response generation by first extracting intention-oriented key clues from the multimodal context through dedicated reasoning, and then employs these clues as auxiliary signals for final knowledge-enhanced textual response generation. 
Notably, we resort to a text-only LLM (\textit{e.g.,} Llama-$3$-$8$B~\cite{dubey2024llama}) rather than a multimodal large language model (MLLM) for textual response generation,  as empirical evidence shows MLLMs of comparable scale exhibit weaker reasoning capabilities~\cite{DBLP:journals/corr/abs-2407-04973}. 
To adapt visual inputs, we convert context images into captions using BLIP~\cite{DBLP:conf/icml/0001LXH22} and feed them together with textual context and filtered related  knowledge into the LLM for response generation. Extensive experiments  on a public dataset verify the superiority of DK2R over existing methods.

Our contribution can be summarized in the following three folds:
\begin{itemize}
\item To the best of our knowledge, we are among the first to integrate both structured attribute knowledge and unstructured review knowledge into the multimodal task-oriented dialog systems.
\item  We propose a novel probe-driven  knowledge type filtering method, which takes advantage of LLMs for identifying the useful type of knowledge in a training-free manner.
\item We propose a two-stage reasoning-enhanced response generation method, which decouples user intention reasoning from response generation for better intention interpretation.  Besides, we  released codes  to benefit the research community\footnote{https://dialogdk2r.wixsite.com/dk2r.}.
\end{itemize}

% The remainder of this paper is organized as follows. Section 2 provides a comprehensive review of related work in the field. Section 3 details our proposed DK2R framework, including its key components and methodology. Extensive experimental results and thorough analyses are presented in Section 4. Finally, Section 5 concludes the paper with key findings and discusses potential directions for future research.

% The remainder of this paper is structured as follows. Section 2 briefly reviews the related work. The proposed DK2R is introduced in Section 3. Section 4 presents the experimental results and analyses, followed by our concluding remarks and future work in Section 5.

\section{Related Work}

\subsection{Task-oriented Dialog Systems }
Recent years have witnessed significant advancements in task-oriented dialog systems, which are designed to accomplish domain-specific tasks ranging from dining reservations to travel arrangements. 
Existing studies~\cite{DBLP:conf/sigir/LiLW21,10.5555/2969033.2969173,DBLP:journals/tkde/LiaoTMYHC22,DBLP:journals/tkde/LiZLC22,DBLP:journals/tois/DengLZDL22}  traditionally employ a pipeline-based  architecture consisting of several interconnected components: \emph{natural language understanding}, \emph{dialog state tracking}, \emph{policy learning}, and \emph{response generation}. Specifically, the initial natural language understanding component  focuses on identifying user intentions, followed by a dialog state tracking component that tracks the 
current state  and populates predefined slots. 
Subsequently, the policy learning  component  determines system actions based on the dialog state representation, while the final response generation component produces outputs either through neural generation techniques or predefined templates.
While pipeline-based approaches have shown promising results, they remain vulnerable to error propagation~\cite{DBLP:conf/acl/KanHLJRY18} and heavy dependence on the sequential  processing components~\cite{DBLP:conf/aaai/ZhangOY20}.

With the advancement of deep learning techniques, researchers have increasingly focused on developing end-to-end task-oriented dialogue systems~\cite{DBLP:conf/aaai/LiYQ22,DBLP:conf/acl/FungWM18,DBLP:journals/tois/WangLL24,DBLP:journals/tois/ZhangHZJCZ19,DBLP:journals/tois/SunGZRCRR24}. 
For example, Li et al~\cite{DBLP:conf/aaai/LiYQ22} investigated emotion recognition in task-oriented dialogue systems by employing BART as the primary architecture, which also incorporates an auxiliary response generation objective to improve contextual comprehension.
Moreover, Madotto et  al.~\cite{DBLP:conf/acl/FungWM18} introduced a memory-to-sequence framework for knowledge-enhanced textual response generation in task-oriented dialogue systems,  which integrates multi-hop attention mechanisms to capture the relation hidden in the knowledge base.
While these approaches have demonstrated promising results, they primarily concentrate on unimodal text-only dialogue systems, overlooking the potential benefits of incorporating visual information. In practical scenarios, both users and conversational agents may need to reference visual content to effectively communicate needs or provide services.

Hence, in recent years, research efforts have increasingly  emphasized multimodal task-oriented dialog systems that extend these capabilities through visual modality integration. 
Along the research line of multimodal task-oriented dialog systems, Saha et al.~\cite{DBLP:conf/aaai/SahaKS18} pioneered the construction of  a large-scale dataset MMD,  and defined two core tasks: textual response generation and image response selection.
Subsequently, a series of studies~\cite{DBLP:conf/sigir/CuiWSHXN19,DBLP:journals/cogsci/Elman90,DBLP:conf/mm/ZhangLGLWN21,DBLP:journals/tip/NieJWWT21,DBLP:conf/acl/ChauhanFEB19} extended this research direction.
For example,  Liao et al.~\cite{DBLP:conf/mm/LiaoM0HC18} introduced a knowledge-aware multimodal dialog model, named KMD, which  integrates the taxonomy-based structured  knowledge to enhance  the visual context representation learning.
Beyond this, Cui et al.~\cite{DBLP:conf/sigir/CuiWSHXN19} designed a user attention-guided multimodal dialog system, named UMD, which first integrates knowledge to enhance the utterance representation and then utilizes the recurrent neural network~\cite{DBLP:journals/cogsci/Elman90} to encode the context and generate responses. Additionally, Zhang et al.~\cite{DBLP:conf/mm/ZhangLGLWN21} proposed  a relational graph-based context-aware question understanding scheme, named TREASURE, which first formulates the dialog context with a graph and sets each utterance as a node.
Thereafter, TREASURE adopts a sparse graph attention network to explore semantic relationships in the context. 
Likewise, Nie et al.~\cite{DBLP:journals/tip/NieJWWT21} constructed a multimodal hierarchical graph that encodes hierarchical semantic relationships among tokens, images, utterances, and turns, and employed the graph attention network for capturing user intention. 
With the rise of Transformer networks, subsequent studies~\cite{DBLP:conf/mm/HeLLCXHY20,DBLP:conf/sigir/ChenSWNC23,DBLP:journals/tomccap/ChenSZWNC25,10.1145/3606368} have adopted Transformer architectures for textual response generation in multimodal task-oriented dialog systems.
For example, He et al.~\cite{DBLP:conf/mm/HeLLCXHY20} utilized the Transformer~\cite{DBLP:conf/nips/VaswaniSPUJGKP17} network to  explore element-level (\textit{e.g.,} words, images) semantic relationships in the given context. 
Besides, Ma et al.~\cite{DBLP:conf/acl/MaL0C22} designed a unified-modal Transformer encoder to project all the multimodal features into a unified semantic space, and a  hierarchical transformer decoder is introduced to generate responses.
In addition, Chen et al.~\cite{DBLP:conf/sigir/ChenSWNC23} adopted the generative pretrained language model BART as the backbone, and integrated both attribute  and relation  knowledge to enhance  textual response generation.

Although the aforementioned studies have achieved significant progress in multimodal task-oriented dialog systems, these methods primarily rely on structured attribute knowledge  while neglecting the importance of unstructured review knowledge and the powerful reasoning capabilities of LLMs. 
To fill this research gap, we integrate dual knowledge (\textit{i.e.,} structured attribute and unstructured review knowledge) with advanced LLMs to enhance  textual response generation in  multimodal task-oriented dialog systems.

\subsection{External Knowledge-Enhanced LLMs}
The emergence of LLMs (\emph{e.g., } ChatGPT\footnote{https://chatgpt.com.} and DeepSeek\footnote{https://www.deepseek.com.}) has brought transformative advances to both natural language processing~\cite{DBLP:conf/acl/ChenGBS023,DBLP:conf/acl/ChungKA23} and computer vision fields~\cite{DBLP:conf/cvpr/TongLZLS0SJ25,DBLP:conf/cvpr/LiMLLZZ25,DBLP:conf/cvpr/Shao0W023}.
The early development of LLMs begins with the \mbox{encoder-decoder} architecture like T$5$~\cite{DBLP:journals/jmlr/RaffelSRLNMZLL20} and BART~\cite{DBLP:conf/acl/LewisLGGMLSZ20}. To be specific, T$5$~\cite{DBLP:journals/jmlr/RaffelSRLNMZLL20} is a unified text-to-text framework by reformulating all NLP tasks into a consistent input-output format, which enables transfer learning through supervised fine-tuning across diverse tasks.
Besides, BART~\cite{DBLP:conf/acl/LewisLGGMLSZ20} adopts a denoising autoencoder architecture that combines bidirectional encoding with autoregressive decoding, particularly effective for text generation and reconstruction tasks through its pre-training objective of reconstructing corrupted text.
Building upon these foundations, GPT-$3$~\cite{DBLP:conf/nips/BrownMRSKDNSSAA20}  eliminates both task templates and fine-tuning, relying instead on in-context learning with raw task descriptions based on the decoder-only architecture.
Furthermore, LLaMA~\cite{DBLP:journals/corr/abs-2302-13971} advances this paradigm as an open-weight foundation model, incorporating architectural improvements like RMSNorm~\cite{DBLP:conf/nips/ZhangS19a} and SwiGLU~\cite{DBLP:journals/corr/abs-2002-05202} activations while maintaining strong few-shot performance across diverse benchmarks.

However, these advances remain fundamentally constrained by the static knowledge acquired during pretraining, limiting their adaptability to specialized domains and evolving real-world information. This has motivated recent efforts~\cite{DBLP:conf/acl/TianGXLHWWW20,DBLP:conf/acl/ZhangHLJSL19,10.1145/3709138} to enhance LLMs through explicit knowledge integration techniques. 
For example, Tian et al.~\cite{DBLP:conf/acl/TianGXLHWWW20} enhanced sentiment analysis performance by integrating domain-specific sentiment words, polarity cues, and aspect-sentiment pairs during pretraining. 
Similarly,  Zhang et al.~\cite{DBLP:conf/acl/ZhangHLJSL19} enriched language representations by incorporating both large-scale textual corpora and structured knowledge graphs into the pretraining framework.
While these approaches have shown promising results, they primarily integrate external knowledge during the pretraining phase, which struggles to adapt to real-world knowledge evolution.
To enable dynamic knowledge integration, researchers have explored leveraging external knowledge to enhance LLMs during the inference phase~\cite{DBLP:conf/nips/LewisPPPKGKLYR020,DBLP:conf/emnlp/WilmotK21,10.1145/3711857}.
For example, Lewis et al.~\cite{DBLP:conf/nips/LewisPPPKGKLYR020} 
designed the retrieval-augmented generation model, which dynamically accesses the Wikipedia\footnote{https://simple.wikipedia.org/wiki/Main\_Page.} knowledge to enhance the generation ability of pretrained language models.
Similarly, Wilmot et al.~\cite{DBLP:conf/emnlp/WilmotK21} 
developed the retrieval mechanism to integrate external knowledge to facilitate inferring salience in long-form stories.
Besides, Wu et al.~\cite{DBLP:conf/emnlp/WuZHMS022} investigated the efficient memory-augmented transformer model that stores external knowledge in a key-value memory, significantly improving performance on knowledge-intensive tasks while maintaining computational efficiency.

Building upon the aforementioned research foundation, we also resorted to utilizing external knowledge to enhance LLMs' response generation capabilities for multimodal task-oriented dialog systems. 
Unlike existing methods, we first extracted context related structured attribute knowledge and unstructured review knowledge, and then employed a probe-driven knowledge filtering mechanism to identify the useful knowledge types.
In this manner, we can achieve the dynamic knowledge type selection, thereby mitigating the issue of irrelevant knowledge integration. 

\section{DK2R}
In this section, we first give problem formulation and then present the three key components of our DK2R, including \textit{context-related dual knowledge extraction}, \textit{probe-driven knowledge type filtering}, and \textit{two-stage reasoning-enhanced response generation}.
To be specific, the context-related dual knowledge extraction component first acquires both structured attribute and unstructured review knowledge relevant to the current  context. Subsequently,  the probe-driven  knowledge type filtering component evaluates the relative utility of each knowledge type, dynamically identifying the useful type of knowledge. Ultimately, the two-stage reasoning-enhanced response generation component first  summarizes \mbox{intention-oriented} key clues, and then conducts the textual response generation.

\subsection{Problem Formulation}
The mathematical notations used in this work are formally defined as follows. Following standard conventions, matrices are denoted by bold uppercase letters (\emph{e.g.,} $\mathbf{X}$), while vectors are represented by bold lowercase letters (\emph{e.g.,} $\mathbf{x}$). Scalar quantities are expressed using non-bold letters (\emph{e.g.,} $x$), and parameters are designated by Greek symbols (\emph{e.g.,} $\gamma$). Unless otherwise specified, all vectors are assumed to be column vectors. 
% A comprehensive summary of the key notations is provided in Table~\ref{tab:notation}.

Suppose we have  $N$  dialogs $\mathcal{D} = \{ (\mathcal{C}_1, R_1), (\mathcal{C}_2,R_2), \cdots, (\mathcal{C}_N, R_N)\}$.
Thereinto, each dialog consists of a multimodal context $\mathcal{C}_i$ and a ground truth response $R_i$.
Each multimodal context $\mathcal{C}_i$ spans multiple utterances, where each utterance contains mandatory text and optional images. To facilitate method description, we aggregate all context tokens and images together, respectively. 
Accordingly,  $\mathcal{C}_i$ can be divided into two parts: (1) $T_i = {[t^i_g]}^{N_T^i}_{g=1}$, the merged textual token sequence with $N_T^i$  tokens, where $t^i_g$ is the $g$-th token; and (2)  $\mathcal{V}_i = \{v^i_j\}^{N_V^i}_{j=1}$, the aggregated image set with $N_V^i$ images, where $v^i_j$ is the $j$-th image in $\mathcal{V}_i$.
Notably, $\mathcal{V}_i = \emptyset$ if there is no image in the context ${\mathcal{C}_i}$.
The ground truth response can be represented as $R_i = {[r^i_n]}^{N_R^i}_{n=1}$ with $N_R^i$ tokens, where $r^i_n$ is the $n$-th token.
Besides, there is an external knowledge base $\mathcal{K} = \{{e_p}\}^{N_K}_{p=1}$, where $e_p$ is the $p$-th knowledge entity  and  $N_K$ is the total number of knowledge entities.
For each  knowledge entity $e_p$, there are two types of knowledge: 1) the structured attribute knowledge $\mathcal{A}_p$, which can be represented as attribute-value pairs (\emph{e.g.,} <domain: food>, <address: Singapore>); and 2) the unstructured review knowledge $\mathcal{U}_p$, \emph{i.e., } the user reviews of the knowledge entity $e_p$.
Meanwhile, for each knowledge entity $e_p$, there are also several images $\mathcal{I}_p$ exhibiting its visual information, facilitating the related knowledge entity identification of a given context based on visual clues. 

Our objective is to learn a mapping function $\mathcal{F}$ with learnable parameters ${\boldsymbol{\Theta}_F}$  that generates contextually appropriate responses given the context $\mathcal{C}_i$ and external knowledge base $\mathcal{K}$ formulated as,
\begin{equation}
    \mathcal{F}(\mathcal{C}_i, \mathcal{K}|\boldsymbol{\Theta}_F)\rightarrow{R_i}.
     \label{eq1}
 \end{equation}

\subsection{Context-Related Dual Knowledge Extraction}
Multimodal task-oriented dialog systems fundamentally depend on external knowledge to generate accurate and informative responses~\cite{10.1145/3474085.3475568}. Unlike existing methods that focus solely on attribute knowledge, we leverage both structured attribute knowledge and unstructured review knowledge. The key lies in identifying context-related knowledge entities, thereby the associated attributes and reviews of these entities form the context-related structured attribute and unstructured review knowledge, respectively.

To ensure a comprehensive extraction of related knowledge entities,
we jointly consider both visual and textual context, following prior work~\cite{DBLP:conf/sigir/ChenSWNC23,10.1145/3606368,DBLP:journals/tomccap/ChenSZWNC25}.  
For textual context, we directly extract the mentioned knowledge
entities based on name matching within the context,  obtaining structured attribute knowledge ${\mathcal{K}^A_t}=\{{\mathcal{A}^1_t}\cup{\mathcal{A}^2_t}\cup \cdots \cup {\mathcal{A}^{N^t_k}_t}\}$ and unstructured review knowledge ${\mathcal{K}^U_t}=\{{\mathcal{U}^1_t}\cup{\mathcal{U}^2_t}\cup \cdots \cup {\mathcal{U}^{N^t_k}_t}\}$, where ${\mathcal{A}^q_t}$ and ${\mathcal{U}^q_t}$ are the structured attribute knowledge and unstructured review  knowledge of  the $q$-th  knowledge entity   related to the textual context, respectively.
 $N^t_k$ is the total number of knowledge entities related to the textual context.
 For visual context $\mathcal{V}_i$ (which may contain multiple images), we retrieve related knowledge entities by processing each context image individually. Using CLIP-pretrained ViT-B/32~\cite{DBLP:conf/iclr/DosovitskiyB0WZ21},  we compute visual cosine similarities between each context image and all images associated with candidate knowledge entities. Since each candidate  entity may have multiple images, we take the highest similarity score between the context image and any of the entity's images as the final matching score. To maintain knowledge quality, we retain only entities whose matching scores exceed a predefined threshold $\theta$ as the related knowledge entities.
This process yields visual-related structured attribute knowledge ${\mathcal{K}^A_v}=\{{\mathcal{A}^1_v}\cup{\mathcal{A}^2_v}\cup \cdots \cup {\mathcal{A}^{N^v_k}_v}\}$ and unstructured review knowledge ${\mathcal{K}^U_v}=\{{\mathcal{U}^1_v}\cup{\mathcal{U}^2_v}\cup \cdots \cup {\mathcal{U}^{N^v_k}_v}\}$, where ${\mathcal{A}^n_v}$ and ${\mathcal{U}^n_v}$ are the structured attribute knowledge and unstructured review  knowledge of  the $n$-th  knowledge entity related to the visual context, respectively.
 $N^v_k$ is the total number of knowledge entities related to the visual context.
By merging the textual and visual context-related knowledge,  we  obtain the final context-related structured attribute knowledge ${\mathcal{K}_A} = \{{\mathcal{K}^A_t}\cup {\mathcal{K}^A_v}\}$ and unstructured review  knowledge ${\mathcal{K}_U} = \{{\mathcal{K}^U_t}\cup {\mathcal{K}^U_v}\}$.

\subsection{Probe-Driven Knowledge Type Filtering}
To effectively address diverse user inquiries, dialog systems often need to leverage different types of knowledge. Intuitively, attribute knowledge helps answer factual queries, while user reviews better address suggestion-based questions. This indicates that directly integrating both knowledge types without differentiation, while relying solely on the response generation backbone to adaptively select relevant knowledge, risks degrading response quality. Therefore, we propose a pre-selection mechanism to identify useful knowledge types (structured or unstructured) for enhancing response generation. Since ground truth annotations for useful knowledge types are unavailable, we introduce a training-free, probe-driven  knowledge type filtering component with powerful LLMs. This component contains two modules: \textit{parallel knowledge-enhanced probe  generation} and \textit{probe-driven knowledge utility assessment}. Thereinto, the former module employs the LLM to generate provisional probe responses for the given  context guided by the extracted structured attribute and unstructured review knowledge, respectively,  while the latter one instructs the LLM to determine whether each type of knowledge is useful by evaluating the  generated probe responses.

\subsubsection{Parallel Knowledge-Enhanced Probe Generation}
In this module, we separately generate the response based on the given multimodal context and each type of knowledge using the LLM. 
Notably, instead of using an advanced MLLM, we opt for the powerful LLM  Llama-$3$-$8$B~\cite{dubey2024llama} due to its superior performance in the textual response generation task~\cite{DBLP:conf/cikm/HuTZLL0L024,DBLP:conf/cikm/00020G24,10.1145/3701551.3703544,10.1145/3627673.3679687}. This process is formulated
as follows, 
\begin{equation}
    \begin{split}
    \begin{cases}
    \mathcal{V}_i^c = {\mathcal{B}_c}(\mathcal{V}_i), \\
    R_i^a={\mathcal{B}_k}(T_i, \mathcal{V}_i^c, \mathcal{K}_A, {P}_{ProvGen}^A), \\
    R_i^u={\mathcal{B}_k}(T_i, \mathcal{V}_i^c, \mathcal{K}_U, {P}_{ProvGen}^U), 
    \end{cases}
    \end{split}
    \label{eq2_generator}
\end{equation}
where $R_i^a$ and $R_i^u$ are the responses generated using structured attribute knowledge ${\mathcal{K}_A}$ and unstructured review knowledge ${\mathcal{K}_U}$, respectively. 
For processing, we concentrate all attribute-value pairs in ${\mathcal{K}_A}$ together into a token sequence, and similarly combine all user reviews in ${\mathcal{K}_U}$.
Notably, rather than utilizing conventional CNN-based visual features of context images $\mathcal{V}_i$, we convert them into captions $\mathcal{V}_i^c$ using BLIP ${{\mathcal{B}_c}(\cdot)}$, as these textual representations better align with the LLM's embedding space and enhance response generation~\cite{10.1145/3606368,DBLP:journals/tomccap/ChenSZWNC25,DBLP:conf/sigir/WenSCWNC24}.
Here,
${\mathcal{B}_k}(\cdot)$ refers to the pretrained Llama-$3$-$8$B model, while
${P}_{ProvGen}^A$ and ${P}_{ProvGen}^U$ are  the prompt templates designed to guide the model in generating provisional probe responses using structured attribute and unstructured review knowledge, respectively.
Taking ${P}_{ProvGen}^A$ as an example, the specific template structure is as follows:
\begin{tcolorbox}[
    breakable, % 允许跨页
    enhanced, % 启用高级功能
    colback=white, % 白色背景
    colframe=black, % 黑色边框
    width=1\linewidth,
    boxrule=0.5pt, % 边框粗细
    arc=0pt, % 直角边框
    left=2pt, right=2pt, top=2pt, bottom=2pt, % 内边距
    before upper={\parindent0pt},
]
{\sethlcolor{yellow!30}\hl{You are a helpful assistant. Based on the given context, you can generate responses with the help of external knowledge. You should only provide the correct response without repeating the context and instruction.}}
{\sethlcolor{red!20}\hl{Context: \{The textual context: ``You can visit warung nasi pariaman! These are some food images from the restaurant. Looks good! Can you send the address to me and check if they accept credit card?''; The caption of visual context: ``a table with plates of food and a cup of coffee a table topped with plates of food and bowls of rice''.\} The external attribute knowledge: ``venuename warung nasi pariaman, venuescore 7.5/10, venueaddress 738 north bridge rd (kg glam conservation area) 198706 Singapore, credit cards no...''}}
\end{tcolorbox}
\noindent
where the yellow highlighting indicates the fixed elements of the prompt template ${P}_{ProvGen}^A$, while the red highlighting marks the variable content that adapts to specific dialog samples.

\subsubsection{Probe-Driven Knowledge Utility Assessment}
In this module, we instruct the pretrained Llama-$3$-$8$B  model to evaluate whether structured attribute and unstructured review knowledge have contributed useful information for response generation, as follows,
\begin{equation}
    \begin{split}
    \begin{cases}
     j_i^a = {\mathcal{B}_j}(T_i, \mathcal{V}_i^c, \mathcal{K}_A, R_i^a, {P}_{KnowAssess}), \\
     j_i^u = {\mathcal{B}_j}(T_i, \mathcal{V}_i^c, \mathcal{K}_U, R_i^u, {P}_{KnowAssess}),
    \end{cases}
    \end{split}
    \label{eq2_judger}
\end{equation}
where $j_i^a\in\{Yes,  No\}$ and $j_i^u\in\{Yes,  No\}$ represent the evaluation results based on the responses generated with structured attribute and unstructured review knowledge, respectively. ${\mathcal{B}_j(\cdot)}$ represents the  Llama-$3$-$8$B  model.
${P}_{KnowAssess}$ is the prompt template for knowledge utility assessment, as shown below,
\begin{tcolorbox}[
    breakable, % 允许跨页
    enhanced, % 启用高级功能
    colback=white, % 白色背景
    colframe=black, % 黑色边框
    width=1\linewidth,
    boxrule=0.5pt, % 边框粗细
    arc=0pt, % 直角边框
    left=2pt, right=2pt, top=2pt, bottom=2pt, % 内边距
    before upper={\parindent0pt}, % 取消首行缩进
]
{\sethlcolor{yellow!30}\hl{You are a knowledge evaluator that can judge if the external knowledge is useful for responding the context. Given a retrieved knowledge, a context, and a response (referred to as the LLM response) generated by an LLM that integrates the retrieved knowledge, you should determine whether the knowledge provides specific information that directly contributes to the LLM response generation; If the information in the knowledge does not help the LLM response generation, you should point it out with evidence. You should respond with ``Yes'' or ``No'' with evidence of your judgment, where ``No'' signifies that the knowledge is not useful.}}
{\sethlcolor{red!20}\hl{The retrieved knowledge: ``venuename Strangers Reunion, venuescore 7.5/10, restroom yes, wi-fi no...''. Context: \{The textual context: ``Okay, I recommend going to strangers reunion. It is a cafe located in the central region. They sell many different brunch food like waffles, as seen in this picture and coffee! I see, can I check if the cafe has a restroom?''; The caption of visual context: ``a waffle with ice cream and fruit on it''.\} The LLM response: ``According to the information, the strangers' reunion cafe does have a restroom.''}}
\end{tcolorbox}

In this way, we can  determine the useful knowledge type based on values of $j_i^a$ and $j_i^u$, and thus obtain the final informative knowledge $\mathcal{K}_F$. Specifically,  if $j_i^a=Yes$,  we retain the structured attribute knowledge $\mathcal{K}_A$; otherwise, we discard it;
Similarly, if $j_i^u=Yes$, we preserve the unstructured review knowledge $\mathcal{K}_U$; otherwise, we exclude it. 
 If both structured attribute and unstructured review knowledge are judged as useful (\textit{i.e., } $j_i^a=j_i^u=Yes$), we combine them as $\mathcal{K}_F = \mathcal{K}_A \cup \mathcal{K}_U$.

\subsection{Two-Stage Reasoning-Enhanced Response Generation}
Once the relevant knowledge type is identified, a naive approach would be to directly concatenate the dialog context and selected knowledge into a powerful LLM (\textit{e.g.,} Llama-$3$-$8$B) through instruction tuning for response generation. While straightforward, the LLM must simultaneously process multimodal context (including textual dialog history and visual captions) and external knowledge (\textit{e.g.,} structured attributes and unstructured reviews), creating a complex reasoning burden. Notably, while the external knowledge primarily supports response formulation, it provides limited value for understanding the underlying user intention. This potentially leads to misinterpretation of core user needs and consequently, suboptimal responses.

\begin{algorithm}[!t]
    \caption{Dual Knowledge-Enhanced Two-Stage Reasoner.}
      \label{alg:Framework}
    \begin{flushleft}
    \hspace*{0.02in} {\bf Input:}
    $\mathcal{D}$, $\mathcal{K}$.\\
    % , $\mathbf{A}$, $\lambda$, $\tau$, $\varphi$.\\
    \hspace*{0.02in} {\bf Output:} Generate the textual response $\tilde{R}$.
      \end{flushleft}
 \begin{algorithmic}[1]
    \Repeat:
    \State Draw $\mathcal{C}_i$ from $\mathcal{D}$.\
    \State Extract the context-related structured attribute knowledge $\mathcal{K}_A$ from $\mathcal{K}$.\
    \State Extract the context-related unstructured review knowledge $\mathcal{K}_U$  from $\mathcal{K}$.\
    \State  Generate the provisional probe responses $R_i^a$ and $R_i^u$ using $\mathcal{K}_A$ and $\mathcal{K}_U$ according to Eqn.($\ref{eq2_generator}$), respectively.\ 
    \State Evaluate the knowledge utility based on the generated provisional responses $R_i^a$ and $R_i^u$ according to  Eqn.($\ref{eq2_judger}$).
    \State Conduct the intention-oriented key clues reasoning  according to  Eqn.($\ref{eq2_finev}$).\
    \State Generate the textual response $\tilde{R}_i$ according to  Eqn.($\ref{eq2_resp}$).\ 
    \State Update $\boldsymbol{\Theta}_F$.\
    \Until: Objective value converges.
 \end{algorithmic}
 \label{algor_all}
 % \vspace{-1em}
 \end{algorithm}

To address these limitations,  we design a two-stage reasoning-enhanced response generation framework. In the first stage, we utilize a powerful  MLLM to perform intention-aware reasoning and summarize key cues from the multimodal context.
In the second stage, we perform the key cues-aware response generation, where the summarized key cues from the multimodal context are used as auxiliary signals for enhancing the following response generation.   
These auxiliary cues act as a ``reasoning shortcut'' that helps the LLM focus on the most relevant aspects of the complex input space.

\textbf{Intention-Oriented Key Clues Reasoning.}
Given the significant advancements achieved by the MLLM xGen-MM~\cite{DBLP:journals/corr/abs-2408-08872} in the multimodal understanding task~\cite{DBLP:conf/mm/ChenDG0W24,DBLP:conf/mm/JiangZZWC024}, we adopt this model to summarize key cues within the multimodal context.
Specifically, we guide the xGen-MM model to summarize the user needs and keywords conveyed by the multimodal context as follows,
\begin{equation}
    \mathcal{Q}_i={\mathcal{B}_m}(T_i, \mathcal{V}_i,  {P}_{ClueGen}),
     \label{eq2_finev} 
 \end{equation}
where $\mathcal{Q}_i$ denotes the generated key clues, and  ${\mathcal{B}_m}(\cdot)$ refers to the MLLM xGen-MM.
$T_i$ and $\mathcal{V}_i$ are the given textual context and visual context, which are merged into a token sequence. ${P}_{ClueGen}$ is the prompt template used for guiding the model in summarizing key clues, structured as follows.
\begin{tcolorbox}[
    breakable,
    enhanced,
    colback=white,
    colframe=black, 
    width=1\linewidth,
    boxrule=0.5pt, 
    arc=0pt,
    left=2pt, right=2pt, top=2pt, bottom=2pt, 
    before upper={\parindent0pt}, 
]
{\sethlcolor{yellow!30}\hl{You are a helpful assistant. Please extract the user need and three-to-five keywords based on the following information.}}
{\sethlcolor{red!20}\hl{Textual context: ``You can visit loof! These are some pictures from the bar. That's what I want! Any tips when visiting the place? Visual context: <image>''}}
\end{tcolorbox}

\textbf{Key Clues-Aware Response Generation.} Ultimately, we utilize the powerful LLM Llama-$3$-$8$B to generate the textual response based on the  textual context, the captions of the visual context,  extracted key cues, and the filtered relevant knowledge.
In fact, a straightforward way to leverage Llama-$3$-$8$B is by utilizing its native inference capabilities  without additional fine-tuning.
However, since Llama-$3$-$8$B  is pretrained on general-domain text corpora, it lacks the specialized knowledge required for multimodal  task-oriented dialog systems.
Hence, we finetune Llama-$3$-$8$B with the Low-Rank Adaptation (LoRA)~\cite{DBLP:conf/iclr/HuSWALWWC22} technique to effectively adapt it for effective textual response generation in multimodal task-oriented  dialog systems as follows,
\begin{equation}
    \tilde{R}_i={\mathcal{B}_g}(T_i,  \mathcal{V}_i^c, \mathcal{Q}_i, \mathcal{K}_F, {P}_{RespGen}),
     \label{eq2_resp} 
\end{equation}
where $\tilde{R}_i$ is the generated response for the context $\mathcal{C}_i$, ${\mathcal{B}_g}(\cdot)$ denotes the LLM Llama-$3$-$8$B.
${P}_{RespGen}$ refers to the prompt template  for the textual response generation task, as follows.
\begin{tcolorbox}[
    breakable,
    enhanced,
    colback=white,
    colframe=black,
    width=1\linewidth,
    boxrule=0.5pt,
    arc=0pt,
    left=2pt, right=2pt, top=2pt, bottom=2pt,
    before upper={\parindent0pt}
]
{\sethlcolor{yellow!30}\hl{You are a helpful assistant. Please think and generate the response based on the given context, the pre-extracted context key clues, and related knowledge. Please prioritize using related knowledge to generate responses. If unable to answer, maintain critical thinking and use your own knowledge to generate responses. Furthermore, please do not rely solely on the pre-extracted context key clues, as the provided context key clues may not always be effective.}}
{\sethlcolor{red!20}\hl{The textual context: ``Ok, how about wingstop? The chicken wings there are great. Sure, do you have their phone number?'' The pre-extracted context elements: ``need: asking for wingstop's phone number keywords: wingstop, chicken wings, phone number, ordering food, location''. Context-related knowledge: ``venuename Wingstop, venuescore 7.2/10, telephone +65 6844 9200, dining options delivery...''}}
\end{tcolorbox}

\begin{figure*}[!t]
    \centering
    \includegraphics[scale=0.46]{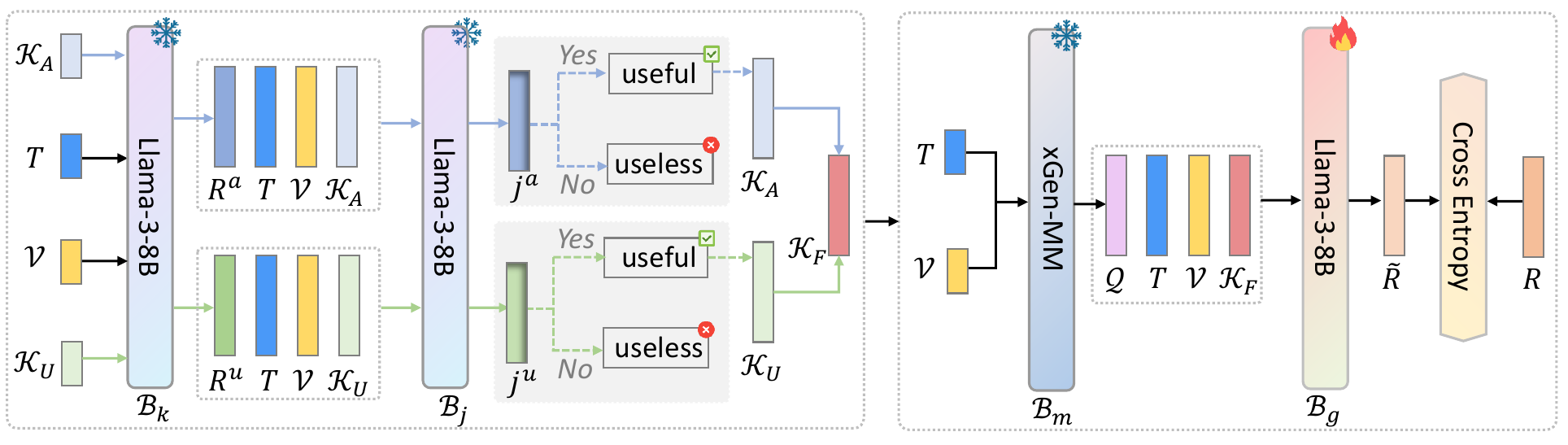}
    \caption{Workflow of the proposed DK2R.}
    % \vspace{-1em}
    \label{Workflow_figure}
\end{figure*}

In particular, considering the limitations of external knowledge, the prompt explicitly states: ``Please prioritize using related knowledge to generate responses. If unable to answer, maintain critical thinking and use your own knowledge to generate responses.''
This strategy aims to fully leverage the implicit knowledge within the LLM, harnessing its powerful logical reasoning capabilities to enhance the quality of textual response generation. 
Additionally, to address the potential inaccuracy of pre-extracted key cues, the prompt includes the following instruction: ``Furthermore, do not rely solely on the pre-extracted contextual key cues, as the provided contextual key cues are not always effective.'' This design improves the robustness and reliability of response generation by preventing over-reliance on potentially inaccurate key clues. 
Notably,   Algorithm~\ref{algor_all} exhibits the procedure, and  Figure~\ref{Workflow_figure} shows the workflow of DK2R.

\section{Experiment}
In this section, we first introduce the dataset and experimental settings, and then provide the experiment results to answer the following research questions:

\begin{itemize}
    \item \textbf{RQ1}: Does DK2R surpass state-of-the-art models?
    \item \textbf{RQ2}: How does  dual knowledge integration contribute to DK2R?
    \item \textbf{RQ3}: How does the reasoning-enhanced response generation affect the performance of DK2R?
    \item \textbf{RQ4}: What is the generalization capability of DK2R when applied to MLLM backbones?
\end{itemize}

\subsection{Dataset}
In the domain of multimodal task-oriented dialogue systems, two publicly available datasets have emerged as standard benchmarks: MMD~\cite{DBLP:conf/aaai/SahaKS18} and MMConv~\cite{DBLP:conf/sigir/LiaoLZHC21}.
While both datasets are widely adopted, MMD is limited to structured attribute knowledge and does not facilitate the integration of unstructured review-based knowledge. Hence, for model evaluation, we employed the public  dataset MMConv, which possesses  both structured attribute knowledge and unstructured review knowledge, facilitating comprehensive knowledge integration.

As demonstrated in Table~\ref{tab:dataset}, the MMConv dataset comprises 5,106  dialogues spanning five distinct domains: \textit{food}, \textit{hotels}, \textit{nightlife}, \textit{shopping mall}, and \textit{sightseeing}. 
The dataset includes 751 unimodal (\emph{i.e.,} only involving text) dialogues and 4,355 multimodal (\emph{i.e., }  involving both text and image) dialogues.
The unimodal dialogues average 7.1 turns in length, while the multimodal dialogues average 7.9 turns, resulting in a total of 39,759 dialogue turns across the dataset.
The dataset includes 113,953 images, with 10,224 images being directly referenced within dialogues. These referenced images appear in 6,858 distinct dialogue turns, and the average number of images in each dialogue is 2.
Additionally, there is an external knowledge base including  1,771 knowledge entities, each equipped with corresponding structured attribute knowledge  and unstructured review knowledge. 
Thereinto, each knowledge entity exhibits  an average of $13.7$ attributes and $24.2$  reviews.
Each knowledge entity is associated with  some images, where the average number of images is $64.3$.

We adhered to the original data split of MMConv, utilizing  3,500 dialogues for training, 606 for validation, and 1,000 for testing. 
Following prior studies~\cite{DBLP:conf/mm/LiaoM0HC18,10.1145/3343031.3350923}, we treated each system utterance as the target response, using the preceding two dialogue turns as context.

\begin{table}[!t]
  \centering
  \caption{Details of the MMConv dataset.}
  % \vspace{-1em}
  \setlength{\tabcolsep}{8mm}{
  \begin{tabular}{l|l}
      \hline
      Entry & Number \\
      % \hline
      \hline
      % $\mathcal{D}$ & Dialog pairs for the model training. \\
      % \hline
      \#Total dialogues & 5,106 \\
      \hline
      \#Dialogue turns & 39,759 \\
      \hline
      \#Unimodal (only involving text) dialogues & 751 \\
      \hline 
      \#Average turns of unimodal dialogues  & 7.1 \\
      \hline       
      \#Multimodal (involving both text and image)  dialogues & 4,355 \\
      \hline 
      \#Average turns of multimodal dialogues  & 7.9 \\      
      \hline 
      \#Total images & 113,953 \\
      \hline  
      \#Average number of images in each dialogue & 2 \\
      \hline       
      \#Knowledge base entities & 1,771 \\
      \hline    
      \#Average number of images for each knowledge entity & 64.3 \\
      \hline          
      \#Average number of structured attributes for each knowledge entity & 13.7 \\
      \hline   
      \#Average number of unstructured reviews for each knowledge entity & 24.2 \\
      \hline         
  \end{tabular}}
  % \vspace{-0.5em}
  \label{tab:dataset}
\end{table}

\subsection{Experimental Setting}
We performed 5 epochs of LoRA fine-tuning on the training set, reporting performance on the test set. Additionally, the proposed DK2R is implemented using SWIFT\footnote{https://github.com/modelscope/ms-swift.} (Scalable lightWeight Infrastructure for Fine-Tuning), a lightweight training and inference toolkit. 
$\theta$ is set as $0.1$.
All experiments are conducted on a server equipped with 8 NVIDIA A100 GPUs (40GB memory each).
As for the quantitative evaluation, in line with existing research, we adopted BLEU-$N$ (where $N$ ranges from 1 to 4) and Nist~\cite{10.5555/1289189.1289273} as evaluation metrics. 
Thereinto,  BLEU-$N$ evaluates $n$-gram overlap between generated responses and ground truth responses. 
As an enhanced variant of BLEU-$N$,  Nist employs an \mbox{information-theoretic} weighting mechanism that preferentially weights low-frequency but linguistically significant $N$-grams.
These automated measures provide a quantitative assessment of response quality, where higher values indicate greater alignment with ground truth responses across both lexical and compositional linguistic dimensions.
To complement these automated measures, we performed qualitative analysis assessing four critical aspects of response quality:  Fluency (grammatical correctness and naturalness), Informativeness (content depth and completeness), Relevance (contextual appropriateness), and Logical Consistency (coherence with dialogue history).

\subsection{On Model Comparison}
For a comprehensive comparison, we performed both quantitative and qualitative comparisons.

\textbf{Quantitative Comparison.} To evaluate the effectiveness of the proposed DK2R, we compared it against baselines from four categories: traditional train-from-scratch methods (TFG), generative pretrained language model based methods (GPLM), multimodal large language model-based
 methods (MLLM), and large language model-based methods (LLM).
    \begin{itemize}
    
    \item \textbf{MHRED}~\cite{DBLP:conf/aaai/SahaKS18} consists of a hierarchical context encoder and a decoder, where the former adopts the Gated Recurrent Units (GRU)~\cite{DBLP:journals/corr/ChungGCB14} network to encode the multimodal utterance and context, while the latter is devoted to generating responses based on GRU. This baseline generates responses only based on the given context, overlooking external knowledge.

    \item \textbf{LARCH}~\cite{DBLP:journals/tip/NieJWWT21}  introduces a hierarchical graph neural network framework for modeling semantic relationships in multimodal dialogue contexts, where words, images, sentences, utterances, dialogue pairs, and the full session are set as  nodes. In addition, the model incorporates attribute knowledge through the memory network and employs a GRU-based decoder for response generation.

    \item \textbf{MATE}~\cite{DBLP:conf/mm/LiaoM0HC18} leverages the Transformer architecture~\cite{DBLP:journals/corr/VaswaniSPUJGKP17} to model semantic dependencies between textual and visual inputs, and  employs a Transformer-based decoder for  response generation.

    \item \textbf{UMD}~\cite{DBLP:conf/sigir/CuiWSHXN19} follows a standard hierarchical encoder-decoder architecture for textual response generation. Specifically, it incorporates a taxonomy-aware tree encoder to extract attribute-level visual representations, along with a factorized bilinear pooling mechanism to obtain multimodal utterance representations.

    \item \textbf{TREASURE}~\cite{DBLP:conf/mm/ZhangLGLWN21} designs a knowledge-enhanced encoder to leverage attribute knowledge to refine the utterance representation through adaptive attention to attribute-relevant keywords. 
    Subsequently, the model employs a sparse attention network to capture semantic relationships among utterances.

    \item \textbf{DialoGPT}~\cite{zhang-etal-2020-dialogpt} is a Transformer-based textual response generation model pretrained on a large-scale conversational corpus. In this work, we cascaded  the textual context, visual context captions, and relevant structured attribute knowledge as input to generate responses.

    \item \textbf{FLAN-T$5$}~\cite{DBLP:conf/iclr/WeiBZGYLDDL22}  is a generative pretrained language model based on an encoder-decoder Transformer architecture. In this work, we combined textual context, visual context captions, and relevant structured attribute knowledge as input, and then fed them into the model to generate textual responses.
    % \item 

    \item \textbf{DKMD}~\cite{10.1145/3606368} employs the pretrained BART model as its backbone, integrating relevant structured attribute knowledge from both global and local perspectives. It also designs a knowledge-enhanced BART decoder to explicitly facilitate textual response generation.

    \item \textbf{MDS-S$^2$}~\cite{DBLP:conf/sigir/ChenSWNC23} also adopts the pretrained BART model as its backbone, incorporating context-related attribute and relational knowledge. Besides, it introduces the latent semantic regularization to refine the user intention representation and thus enhance textual response generation.

\begin{table}[!t]
    \centering
    \small
    \caption{Performance comparison among DK2R and baselines. The best results are shown  in \textbf{bold}, and the second best are \underline{underlined}. ``TFG'' refers to train-from-scratch methods.} 
    \setlength{\tabcolsep}{6mm}{\begin{tabular}{c|l|rrrrr}
    \hline
        \multicolumn{2}{c|}{Methods} & BLEU-1 & BLEU-2 & BLEU-3 & BLEU-4 & Nist \\ 
        \hline

        \multirow{5}{*}{\rotatebox{90}{{TFG}}} & MHRED & $15.02$ & $6.66$ & $4.24$ & $2.94$ & $0.9529$ \\

        & LARCH & $20.86$ & $11.33$ & $7.58$ & $5.58$ & $1.3400$ \\ 

        & MATE & $30.45$ &$22.06$ & $17.05$ & $13.41$ & $2.3426$ \\ 
        
        & UMD & $31.14$ & $21.87$ & $17.12$ & $13.82$ & $2.5290$ \\ 
        
        & TREASURE & $34.75$ & $24.82$ & $18.67$ & $14.53$ & $2.4398$ \\ 
        \hline

        \multirow{4}{*}{\rotatebox{90}{{GPLM}}}    & DialoGPT & $32.58$ & $23.83$ & $19.22$ & $15.98$ & $2.8182$ \\ 
        
        & Flan-T$5$ & $36.01$ & $27.99$ & $23.38$ & $20.03$ & $3.5890$ \\
        
        & {DKMD} & ${39.59}$ & $31.95$ & ${27.26}$ & ${23.72}$ & ${4.0004}$ \\ 
        
        & MDS-S$^2$ & $41.40$ & $32.91$ & $27.74$ & $23.89$ & $4.2142$ \\ 
        \hline 

        % \rowcolor{gray!5}\multicolumn{6}{c}{\textit{\textcolor{gray}{Multimodal Large Language Model-Based Methods}}}  \\         
        \multirow{3}{*}{\rotatebox{90}{{MLLM}}}& Yi-VL-$6$B & $40.55$ & $34.07$ & $30.21$ & $27.46$ & $3.0728$ \\

        & Qwen$2.5$-VL-$7$B & $42.14$ & $35.30$ & $31.27$ & $28.43$ & $3.5536$ \\
        
        % & InternVL$2$\_$5$-$8$B & $45.18$ & $38.94$ & $34.76$ & $31.55$ & $3.8069$ \\ 
        
        & LLaVA-$1.6$-$7$B & ${45.67}$ & $38.04$ & ${33.63}$ & ${30.41}$ & ${5.0993}$ \\ \hline % LLaVA1_6-vicuna-7b-instruct
        % & LLaVA-$1.6$-$7$B-U & ${47.69}$ & $39.93$ & ${35.44}$ & ${32.15}$ & ${5.4309}$ \\ \hline % LLaVA1_6-vicuna-7b-instruct
        % & {LLaVA-$1.6$-$7$B$+$} & {$46.53$} & {$38.70$} & {$34.18$} & {$30.89$} & {$5.2432$} \\ \hline

        % \rowcolor{gray!5}\multicolumn{6}{c}{\textit{\textcolor{gray}{Large Language Model-Based Methods}}}  \\            
        \multirow{6}{*}{\rotatebox{90}{{LLM}}}&GPT-3.5 Turbo & $20.01$ & $11.81$ & $7.68$ & $5.24$ & $1.6532$ \\ 

        & GPT-4 Turbo & $25.68$ & $17.21$ &$12.68$ & $9.72$ & $2.3841$ \\

        & Qwen$2.5$-$7$B & $37.08$ & $30.11$ & $26.05$ & $23.04$ & $3.1506$ \\
        
         % Llama-$2$-$7$B & $43.64$ & $36.02$ & $31.70$ & $28.60$ & $4.7854$ \\ 
         
        & DeepSeek-llm-$7$B & $44.04$ & $38.38$ & $34.85$ & \underline{$32.13$} & $3.5179$ \\         
        & {Llama-$3$-$8$B} & \underline{${47.48}$} & \underline{$39.48$} & $\underline{{34.85}}$ & ${31.47}$ & \underline{${5.3308}$} \\ 
        
        % & \rowcolor{gray!15} \textcolor{defblue}{DK2R} & \textcolor{defblue}{$49.63$} & \textcolor{defblue}{$41.71$} & \textcolor{defblue}{$37.10$} & \textcolor{defblue}{$33.71$} & \textcolor{defblue}{$5.6856$} \\ 
        % \hline     
        &   \textbf{DK2R} & $\textbf{49.63}$ & $\textbf{41.71}$ & $\textbf{37.10}$ & $\textbf{33.71}$ & $\textbf{5.6856}$ \\ \hline         
    \end{tabular}}
    \label{tab_rq1}
    % \vspace{-1.2em}
\end{table}

    \item \textbf{GPT-3.5 Turbo}~\cite{DBLP:conf/nips/BrownMRSKDNSSAA20} is a text-based large language model, which belongs to the GPT series architecture and is specifically fine-tuned for conversational AI applications~\cite{DBLP:conf/nips/BrownMRSKDNSSAA20}. In our experimental framework, we implemented this model through its official API interface\footnote{https://platform.openai.com/docs/models/gpt-3-5-turbo.}, employing in-context learning methodologies to effectively leverage its capabilities~\cite{DBLP:conf/mm/QuW0NC23}.

    \item \textbf{GPT-4 Turbo}~\cite{DBLP:journals/corr/abs-2303-08774} represents an enhanced iteration of the GPT-4  architecture, functioning as a multimodal large language model that has been specifically optimized for conversational AI applications.  In our experimental framework, we utilized this model through its  API interface\footnote{https://platform.openai.com/docs/models/gpt-4-turbo-and-gpt-4.}, employing in-context learning techniques to fully realize its functional potential.

    \item \textbf{Yi-VL-$6$B}~\cite{DBLP:journals/corr/abs-2403-04652} is a multimodal large language model with $6$ billion parameters. In this work, we integrated  textual context, original visual images, and structured attribute knowledge as  input. We also employed the LoRA technique to fine-tune the model.

    \item \textbf{Qwen$2.5$-VL-$7$B}~\cite{bai2025qwen25vltechnicalreport} is a  multimodal generative language model with $7$ billion parameters, which is the latest advancement in the Qwen series of large language models with vision-language capabilities. Similarly, we combined textual context, captions of visual context, and relevant structured attribute knowledge as input, and adopted the LoRA technique to finetune the model.

    % \item  

    % \item \textbf{InternVL$2$\_$5$-$8$B}~\cite{DBLP:journals/corr/abs-2401-02954} is a multimodal large language model comprising 8 billion parameters. Similarly, we combined textual context, captions of visual context, and relevant structured attribute knowledge as input, and adopted the LoRA technique to finetune the model.
    % \item 

    \item \textbf{LLaVA-$1.6$-$7$B}~\cite{DBLP:conf/nips/LiuLWL23a} is a $7$-billion-parameter multimodal large language model pretrained on large-scale image-text datasets. We also fine-tuned the model using  the LoRA technique.
    % \item \textbf{LLaVA-$1.6$-$7$B$+$} is a extended version of LLaVA-$1.6$-$7$B, which implements the proposed DK2R based on  LLaVA-$1.6$-$7$B.
    % Namely, we adopted LLaVA-$1.6$-$7$B as the backbone of DK2R.

    \item  \textbf{Qwen2.5-7B}~\cite{qwen2025qwen25technicalreport} is a 7-billion-parameter generative language model. In this work, we concatenated textual context,  captions of visual context, and relevant knowledge as input. We also adopted the LoRA technique to fine-tune the model.

    \item \textbf{DeepSeek-llm-$7$B}~\cite{DBLP:journals/corr/abs-2401-02954} is a language model comprising 7 billion parameters, pretrained on a vast dataset of 2 trillion tokens. Similarly, we combined textual context, captions of visual context, and relevant structured  knowledge as input, and adopted the LoRA technique to finetune the model.

    \item \textbf{Llama-$3$-$8$B}~\cite{dubey2024llama} is a generative large language model with  $8$ billion parameters. 
    Similarly, we concentrated on the textual context, captions of visual context, and related structured attribute knowledge as the input. We also fine-tuned the model using the LoRA technique.

\end{itemize}

    % \begin{table}[!t]
    % \centering
    %     \caption{Human evaluation on responses generated by DK2R and Llama-$3$-$8$B across  four evaluation factors.}
    %     \setlength{\tabcolsep}{12mm}{\begin{tabular}{l|ccc}
    % \hline
    %  Evaluation Factors &  Win &  Loss &  Tie  \\ \hline
    %   Fluency &  $76.00\%$  &  $23.33\%$ &  $0.67\%$ \\ 
    %  Informativeness  & $80.00\%$   & $16.67\%$  & $3.33\%$ \\  
    %   Relevance  & $74.00\%$   & $22.00\%$ & $4.00\%$    \\  
    %  Logical Consistency &  $74.67\%$ &  $23.33\%$ &  $2.00\%$   \\ \hline
    % \end{tabular}}
    % \label{rq1_human}
    % % \vspace{-1em}
    % \end{table}

\begin{figure}[!t]
    \centering
 \subfigure[Fluency.]{
  \includegraphics[scale=0.23]{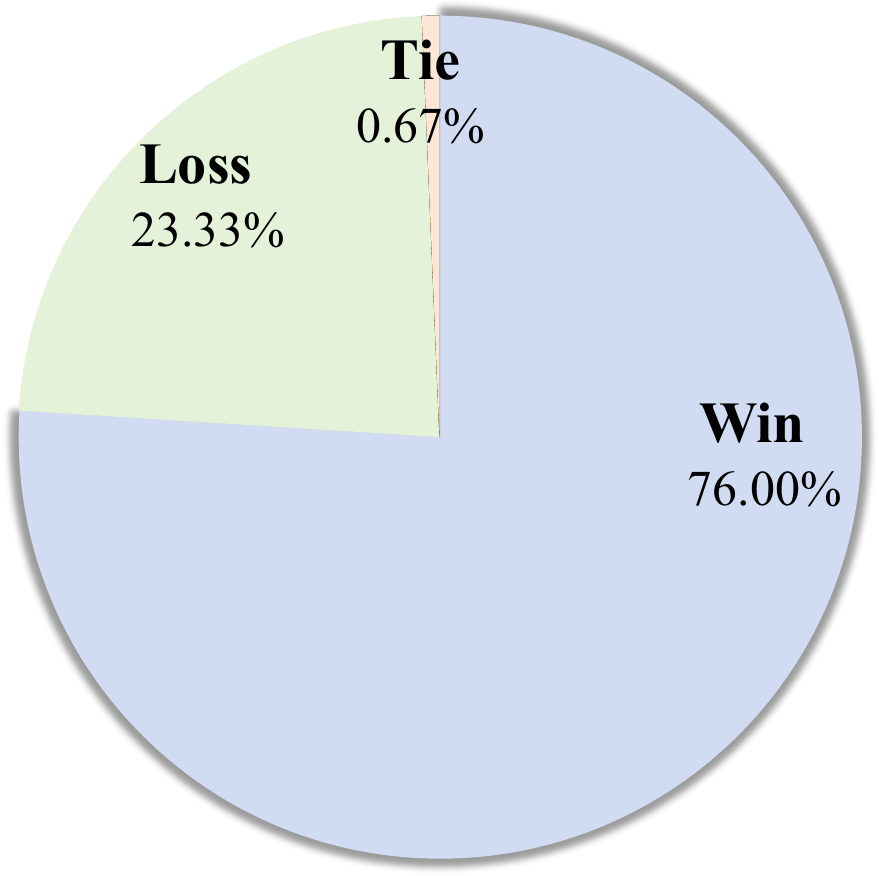}
  }
 \subfigure[Informativeness.]{
  \includegraphics[scale=0.23]{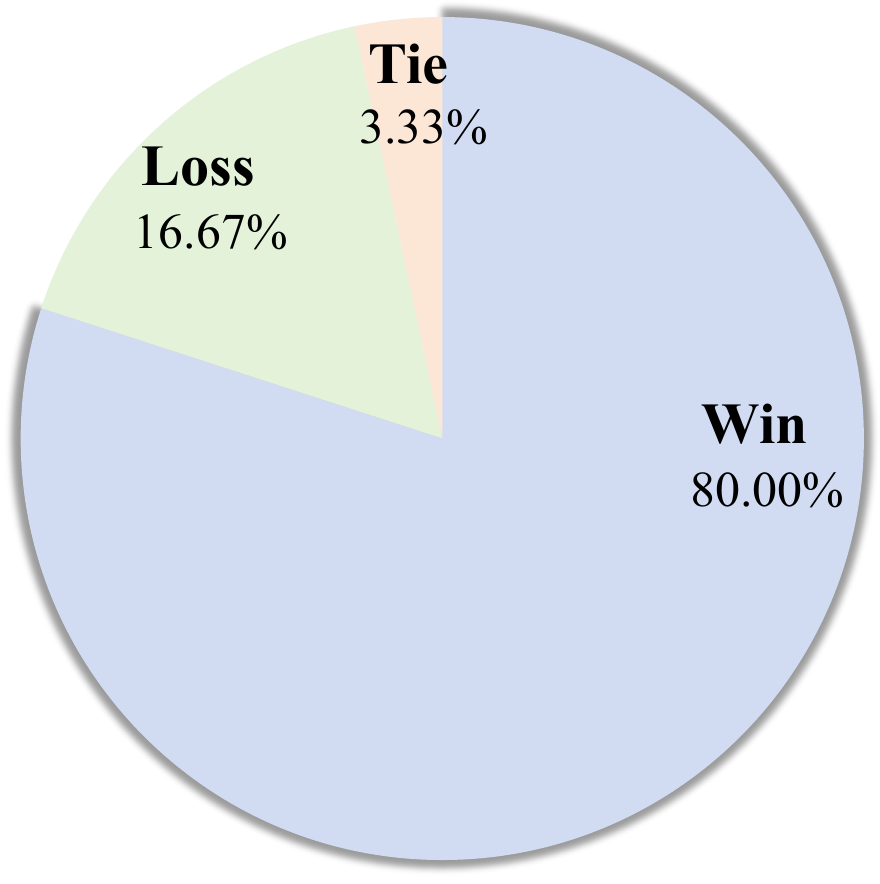}
  }
  \subfigure[Relevance.]{
  \includegraphics[scale=0.23]{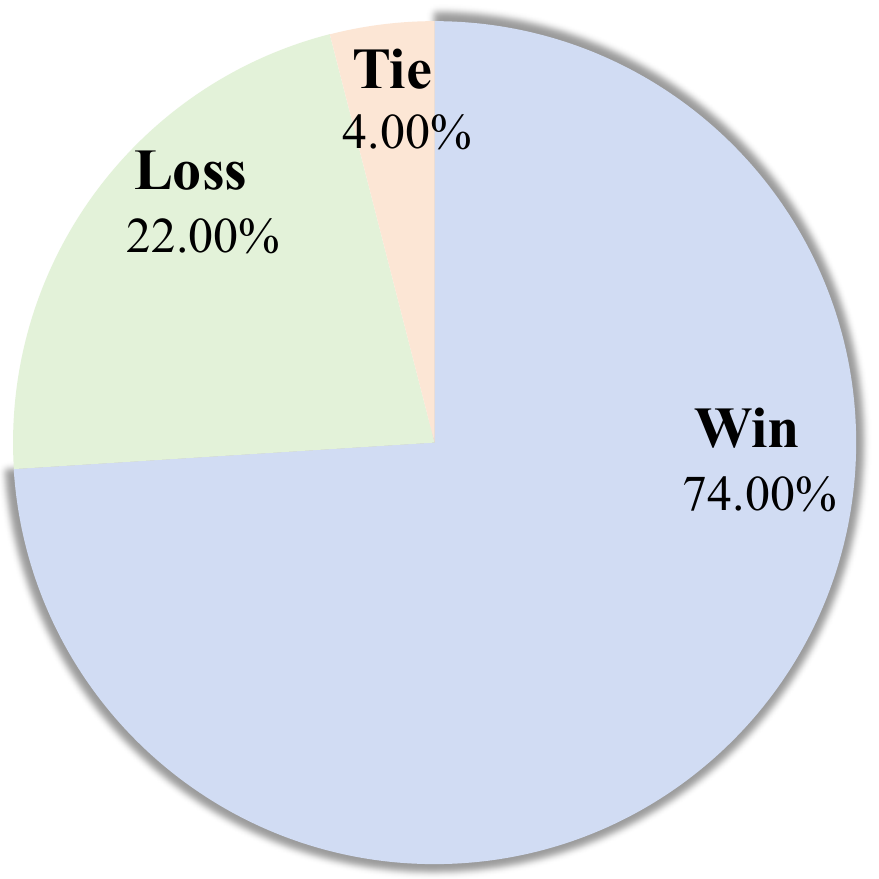}
  }
  \subfigure[Logical Consistency.]{
  \includegraphics[scale=0.23]{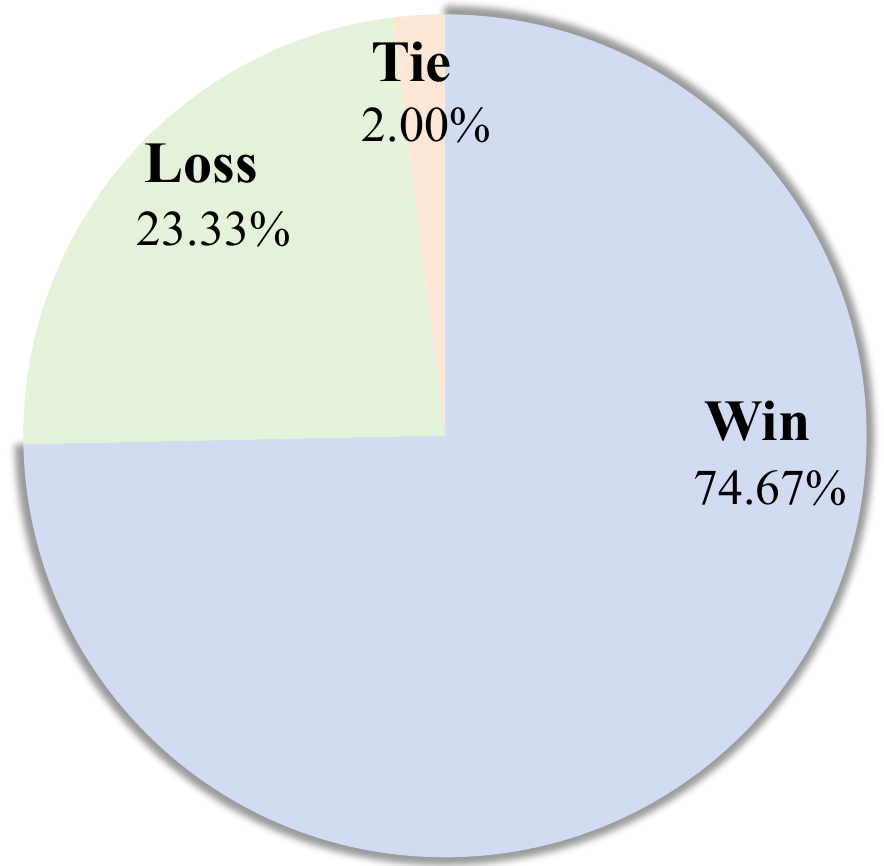}
  }  
  \caption{Human evaluation on responses generated by DK2R and Llama-$3$-$8$B across  four evaluation factors. }
    % \vspace{-1em}
    \label{rq1_human}
\end{figure}

Table~\ref{tab_rq1} exhibits the performance comparison between DK2R and baselines.  
From the table,  we had the following observations. 
\mbox{1) DK2R} demonstrates superiority over  all the baselines, including those LLM- or MLLM-based methods, across all evaluation metrics, exhibiting its superiority. This advantage can be attributed to two key factors: (1) our model integrates unstructured review knowledge, and (2) it decouples user intention understanding from knowledge-enhanced response generation.
2) Llama-$3$-$8$B surpasses all baselines based on MLLMs (\textit{i.e.,} Yi-VL-$6$B, Qwen$2.5$-VL-$7$B,  and  LLaVA-$1.6$-$7$B) and LLMs (\textit{i.e., } GPT-$3.5$ Turbo, GPT-$4$ Turbo, Qwen$2.5$-$7$B and DeepSeek-llm-$7$B), which  highlights  the strong reasoning capability of Llama-$3$-$8$B and confirms the advantage of text-only LLMs over MLLMs in textual response generation. 
3) DK2R achieves better performance than its backbone model Llama-$3$-$8$B, further validating the benefits of incorporating  unstructured review  knowledge and  decoupling user intention understanding and response generation.
And 4) MHRED exhibits the most inferior performance among all baselines. This performance deficiency fundamentally originates from its complete neglect of external knowledge integration, which severely restricts its capacity to generate appropriate responses.

\begin{figure}[!t]
    \centering
 \subfigure[Case 1.]{
  \includegraphics[scale=0.41]{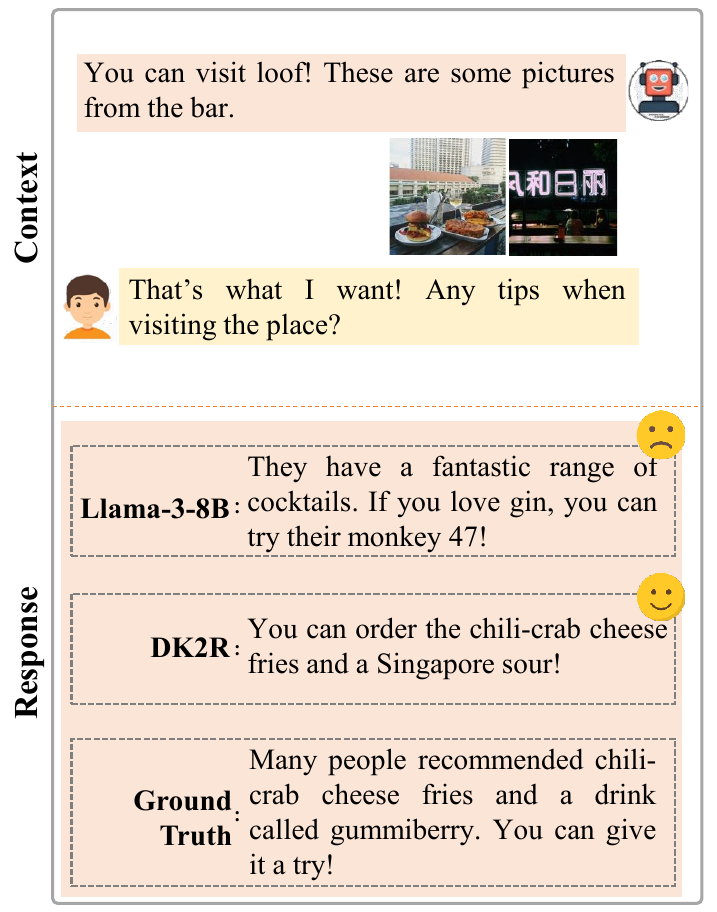}
  }
 \subfigure[Case 2.]{
  \includegraphics[scale=0.41]{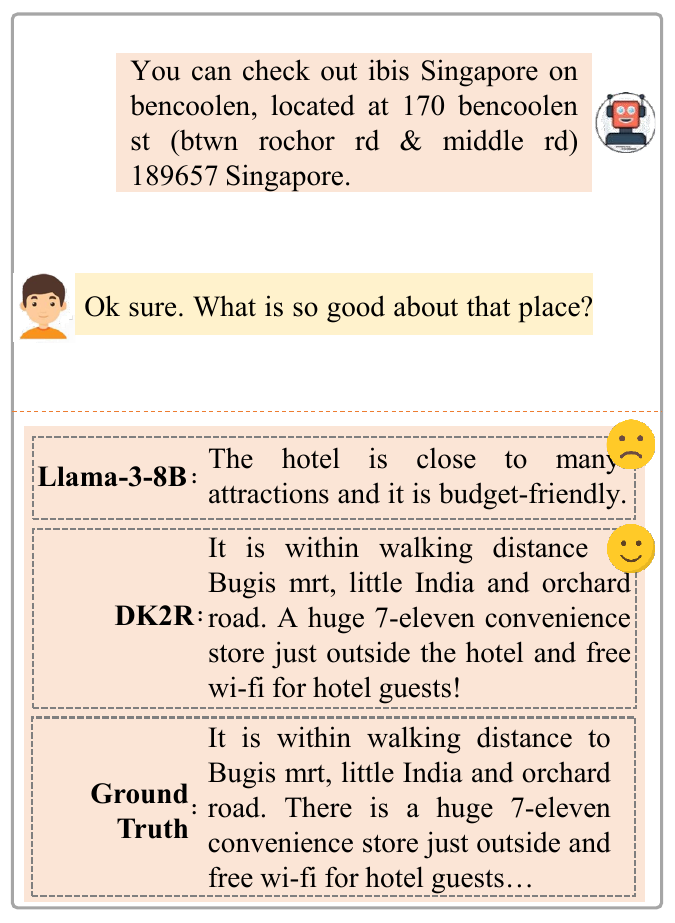}
  }
  \subfigure[Case 3.]{
  \includegraphics[scale=0.41]{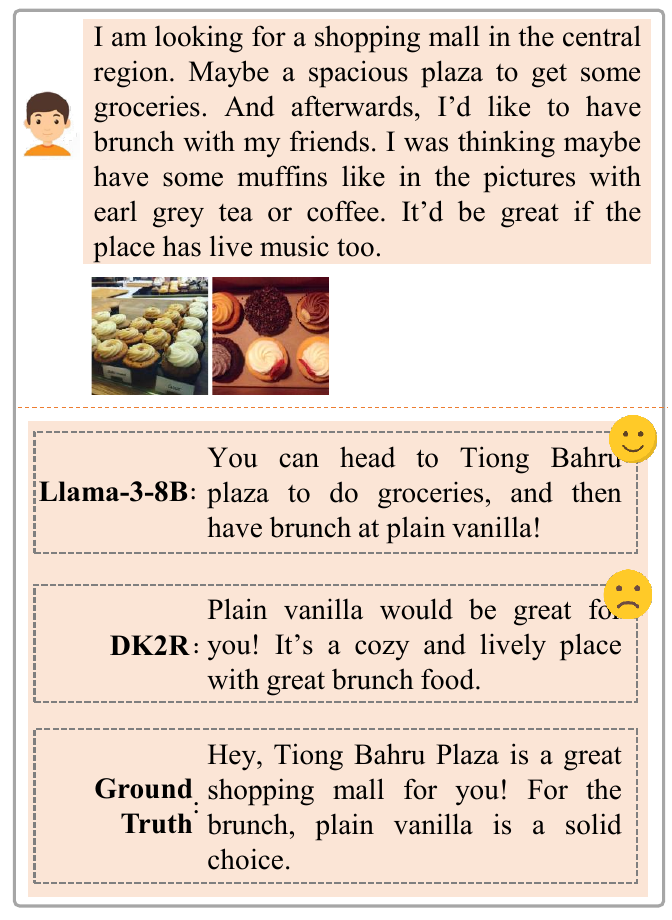}
  }
  \caption{Comparison between our DK2R and Llama-$3$-$8$B on three testing dialog pairs. The smiling emoji indicates the preferred response, while the sad one marks the less desirable one.}
    \vspace{-1em}
    \label{RQ1_Case}
\end{figure}

\textbf{Qualitative Comparison.} To fully assess  the quality of generated responses beyond automated metrics, we  also  conducted the human evaluation comparing DK2R against the best baseline Llama-$3$-$8$B.
Specifically, we first randomly selected $50$ testing samples, and generated responses using both DK2R and Llama-$3$-$8$B.
Thereafter, following prior work ~\cite{DBLP:conf/sigir/CuiWSHXN19,DBLP:journals/tomccap/ChenSZWNC25}, we engaged  three  annotators to independently evaluate  the responses based on four criteria: Fluency,  Informativeness, Relevance, and Logical Consistency. 
To eliminate bias, the order of responses is shuffled randomly.
% Notably, the order of responses is shuffled randomly.
Figure~\ref{rq1_human} summarizes the average judgments across annotators.
Specifically, ``Win'' refers that the response generated by DK2R outperforms  that generated by Llama-$3$-$8$B, while ``Loss'' denotes that Llama-$3$-$8$B is superior.
``Tie'' suggests that which one is better is indistinguishable.
As can be seen, DK2R exhibits consistent superiority across all evaluation dimensions, further validating its effectiveness over  DK2R.
The results show DK2R achieving dominant win rates (74.00\%-80.00\%) over Llama-3-8B in all categories, with particularly strong performance in Informativeness (80.00\% wins). 
The minimal tie rates (0.67\%-4.00\%) further reinforce  the superior generation capability of DK2R, confirming its effectiveness in producing linguistically fluent, contextually relevant, and logically coherent responses.

To intuitively exhibit the superiority of DK2R, we randomly sampled three testing dialog pairs and illustrated the responses generated by DK2R and the best baseline Llama-3-8B in Figure~\ref{RQ1_Case}.
Thereinto, the smiling emoji indicates the preferred response, while the sad one marks the less desirable one.
% the smiling emoji represents a satisfactory response, while the sad one indicates a suboptimal one.
As we can see, for  \textit{Case 1}, which requests tips for visiting a place, our proposed DK2R that incorporates unstructured review knowledge surpasses Llama-$3$-$8$B.
Specifically, Llama-$3$-$8$B recommends ``monkey 47'', while our proposed DK2R suggests ``chili-crab cheese fries and a Singapore sour''.
Upon consulting the external knowledge base, we noticed that these recommendations of ``chili-crab cheese fries'' and  ``Singapore sour'' provided by our DK2R likely originate from the retrieved context-related unstructured review  knowledge. Specifically, there is one review stating: ``Singapore 2019s first stand-alone rooftop bar has recently injected its menu with Southeast Asian flavor. Order the chili-crab cheese fries with a Singapore Sour''.
Similarly, for \textit{Case 2}, DK2R again outperforms the baseline Llama-$3$-$8$B, where
Llama-$3$-$8$B offers a conventional description of Ibis Singapore (\textit{i.e.,} ``...close to many attractions and it is budget-friendly''), whereas DK2R provides more authentic and location-specific suggestions (\textit{i.e.,}``...within walking distance to Bugis mrt, little India and orchard road. A huge 7-eleven...''), benefited from referring to the retrieved unstructured review  knowledge. 
These findings emphasize the effectiveness of the proposed DK2R that incorporates unstructured review knowledge for enhancing textual response generation of multimodal task-oriented dialog systems.
However, for \textit{Case 3}, DK2R underperforms compared to Llama-$3$-$8$B. While both DK2R and Llama-$3$-$8$B correctly recommend ``Plain vanilla'' for brunch, DK2R fails to address the user's primary request for a shopping mall recommendation. This observation may stem from our proposed DK2R overemphasizing   the detailed brunch preferences (\textit{e.g.,} muffins and live music), which could introduce noise in its reasoning about the user's broader intentions. This suggests that DK2R might struggle with balancing multiple user intentions when one is described more vividly than the other.

\begin{table}[!t]
    \centering
    \small
    % \vspace{-1em}
    \caption{Performance comparison between DK2R and its dual  knowledge integration variants.}
    \setlength{\tabcolsep}{7.2mm}{\begin{tabular}{l|rrrrr}
    \hline
        Methods & BLEU-1 & BLEU-2 & BLEU-3 & BLEU-4 & Nist \\ \hline
        w/o-AllK & $35.90$ & $26.75$ & $21.71$ & $18.17$ & $3.3416$ \\ 
         w/o-StrucK & $40.38$ & $31.11$ & $25.85$ & $22.13$ & $4.0301$ \\
       w/o-UnstrucK & $47.47$ & $39.57$ & $35.01$ & $31.66$ & $5.3473$ \\ 
        w/o-KTypeFilter & $48.61$ & $40.78$ & $36.24$ & $32.93$ & $5.5598$ \\ 
        % \rowcolor{gray!15} \textbf{DK2R} & $\textbf{49.63}$ & $\textbf{41.71}$ & $\textbf{37.10}$ & $\textbf{33.71}$ & $\textbf{5.6856}$ \\ 
        \hline   
         \textbf{DK2R} & $\textbf{49.63}$ & $\textbf{41.71}$ & $\textbf{37.10}$ & $\textbf{33.71}$ & $\textbf{5.6856}$ \\ \hline    
    \end{tabular}}
    \label{chap5:tab_rq2}
\end{table}

\subsection{On Dual Knowledge Integration}
To validate the effectiveness of dual  knowledge integration, we designed the following variants.
\begin{itemize}
    \item{\textbf{w/o-AllK}}. To verify the importance of incorporating knowledge, this variant removes all relevant knowledge by eliminating $\mathcal{K}_F$ from Eqn.~(\ref{eq2_finev}). 
    \item {\textbf{w/o-StrucK}}. To verify the importance of structured attribute knowledge, this variant retains only unstructured review knowledge while removing context-related structured attribute knowledge.
    \item {\textbf{w/o-UnstrucK}}. To verify the role of unstructured review knowledge, this variant retains only structured attribute  knowledge while removing context-related unstructured review knowledge.
    \item \mbox{{\textbf{w/o-KTypeFilter}}.} To verify the importance of knowledge type  filtering, this variant removes the probe-driven knowledge type filtering component. Specifically, it directly uses the initially filtered knowledge $[{\mathcal{K}_A}, {\mathcal{K}_U}]$ instead of ${\mathcal{K}_F}$ in Eqn.~(\ref{eq2_finev}).
\end{itemize}

\begin{figure}[!t]
    \centering
    \includegraphics[scale=0.67]{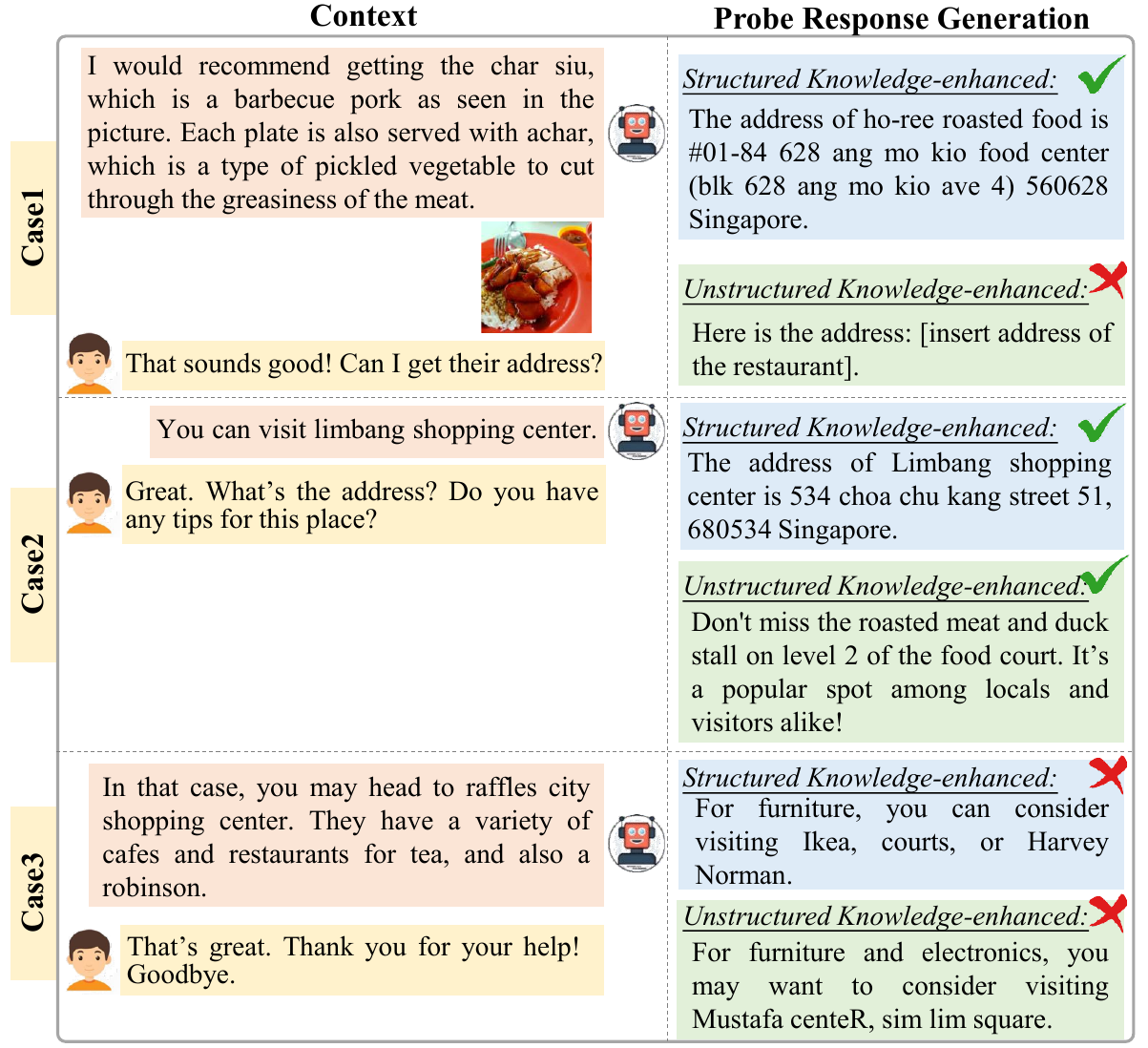}
    \caption{Illustration of the knowledge type filtering. The green tick denotes  the knowledge type is judged as useful, while the red cross represents  the knowledge type is deemed useless.  }
    \vspace{-1em}
    \label{chap5:figure_kplan}
\end{figure}

Table~\ref{chap5:tab_rq2} illustrates the performance comparison between DK2R and its variants. 
The following observations can be made.
1) DK2R outperforms w/o-KTypeFilter, indicating the effectiveness of the probe-driven knowledge type filtering component. 
This experimental phenomenon exhibits that the component effectively  helps identify the type(s) of knowledge to incorporate, thereby reducing interference from irrelevant knowledge.
2) DK2R achieves superior performance compared to w/o-UnstrucK, w/o-StrucK, and w/o-AllK, with w/o-AllK performing the worst results. 
This phenomenon demonstrates that removing either structured attribute or unstructured review knowledge harms model performance, further underscoring the necessity of both structured attribute and unstructured review knowledge in multimodal task-oriented dialogue systems.
% validating the importance of both types of knowledge in multimodal task-oriented dialog systems.
And \mbox{3) w/o-StrucK} performs worse than \mbox{w/o-UnstrucK}, suggesting that structured attribute knowledge  plays a more critical role in  response generation than unstructured review knowledge. A plausible  explanation is that users  in multimodal task-oriented dialogues typically focus more on inquiring about the attribute information of an entity rather than seeking advice.

To provide a more comprehensive analysis of the probe-driven knowledge type filtering component, we randomly sampled three test dialogues, and visualized the parallel \mbox{knowledge-enhanced} provisional probe responses derived by Eqn.~($\ref{eq2_generator}$) alongside their corresponding knowledge utility assessments derived by Eqn.~($\ref{eq2_judger}$) in Figure~\ref{chap5:figure_kplan}.
As can be seen, in \textit{Case 1}, DK2R determines that structured attribute knowledge is useful, while unstructured review knowledge is not. Upon examining the context, we confirm that the user requested address-related information, validating the assessment.
In \textit{Case 2}, DK2R identifies both structured attribute knowledge and  unstructured review knowledge as useful. The context shows that the user inquired about both address details and shopping tips, justifying the model's judgment again.
In \textit{Case 3}, we found that  both  the structured attribute knowledge and  unstructured  review knowledge are deemed useless  by DK2R. 
Checking the context, we found that the user need has been satisfied  and the conversation reached its natural conclusion, suggesting no further knowledge integration is necessary.
These observations provide empirical evidence that the probe-driven knowledge type filtering component effectively identifies and utilizes contextually relevant knowledge while suppressing irrelevant information, thereby optimizing response generation.
% These observations demonstrate that the provisional response-based knowledge type filtering component does identify the useful knowledge type(s) for responding to the given context.

\subsection{On Reasoning-Enhanced Response Generation}
To  validate the effectiveness of reasoning-enhanced response generation, we designed the following variants.
\begin{itemize}
    \item {\textbf{w/o-AllC}}. To verify the importance of  reasoning-enhanced response generation, this variant removes all key clues from  DK2R  by eliminating $Q_i$ from Eqn.~(\ref{eq2_finev}).
\item {\textbf{w/o-Keywords}}. To examine the significance of keywords, this variant discards keywords in key clues, where  $Q_i$ in Eqn.~(\ref{eq2_finev}) consists solely of user needs.
\item {\textbf{w/o-Needs}}. To  validate the importance of user needs in key clues, this variant removes  user needs, where  $Q_i$ in Eqn.~(\ref{eq2_finev}) only contains keywords.
\end{itemize}
\begin{table}[!t]
    \centering
    \small
    \caption{Performance comparison between DK2R and its derivatives of reasoning-enhanced response generation.}
    \setlength{\tabcolsep}{7mm}{\begin{tabular}{l|rrrrr}
    \hline
        Methods & BLEU-1 & BLEU-2 & BLEU-3 & BLEU-4 & Nist \\ \hline
         w/o-AllC & $47.72$ & $40.48$ & $36.25$ & $33.13$ & $5.4550$ \\
        w/o-Need & $48.63$ & $40.71$ & $36.21$ & $32.74$ & $5.5402$ \\ 
        w/o-Keywords & $48.64$ & $40.81$ & $36.27$ & $32.96$ & $5.5592$ \\ 
        % {\color{blue}w/-Yi-VL-$6$B} & {$41.54$} & {$35.25$} & {$31.50$} & {$28.79$} & {$3.2985$} \\             
         % {\color{blue} w/-Qwen$2.5$-VL-$7$B} & {$42.29$} & {$36.55$} & {$33.00$} & {$30.35$} & {$3.5452$} \\        
         % {w/-LLaVA-$1.6$-$7$B} & {$46.53$} & {$38.70$} & {$34.18$} & {$30.89$} & {$5.2432$} \\ \hline
        % \rowcolor{gray!15} \textbf{DK2R} & $\textbf{49.63}$ & $\textbf{41.71}$ & $\textbf{37.10}$ & $\textbf{33.71}$ & $\textbf{5.6856}$ \\ \hline 
        \textbf{DK2R} & $\textbf{49.63}$ & $\textbf{41.71}$ & $\textbf{37.10}$ & $\textbf{33.71}$ & $\textbf{5.6856}$ \\ \hline          
    \end{tabular}}
    \label{chap5:tab_rq3} 
    % \vspace{-1em}
\end{table}

Table~\ref{chap5:tab_rq3} summarizes the performance comparison between DK2R and its ablation variants. The following observations can be made from the table.
\mbox{1) DK2R} demonstrates superior performance over w/o-AllC, confirming the effectiveness of reasoning-enhanced response generation.
This substantiates   that conducting intention-oriented key clues reasoning    
indeed improves the quality of  response generation.
\mbox{2) Both} w/o-Need and w/o-Keywords perform worse than DK2R, which exhibits that both keywords and user needs contribute to the textual response generation. 
And 3)  w/o-Need achieves a worse performance than w/o-Keywords, which shows that user needs play a more critical role than  keywords towards   response generation.
This is likely because the user needs directly capture the intention behind the user  requests, providing more semantically rich guidance than keyword information alone.

\begin{figure}[!b]
    \centering
 \subfigure[Word cloud of user needs.]{
  \includegraphics[scale=0.065]{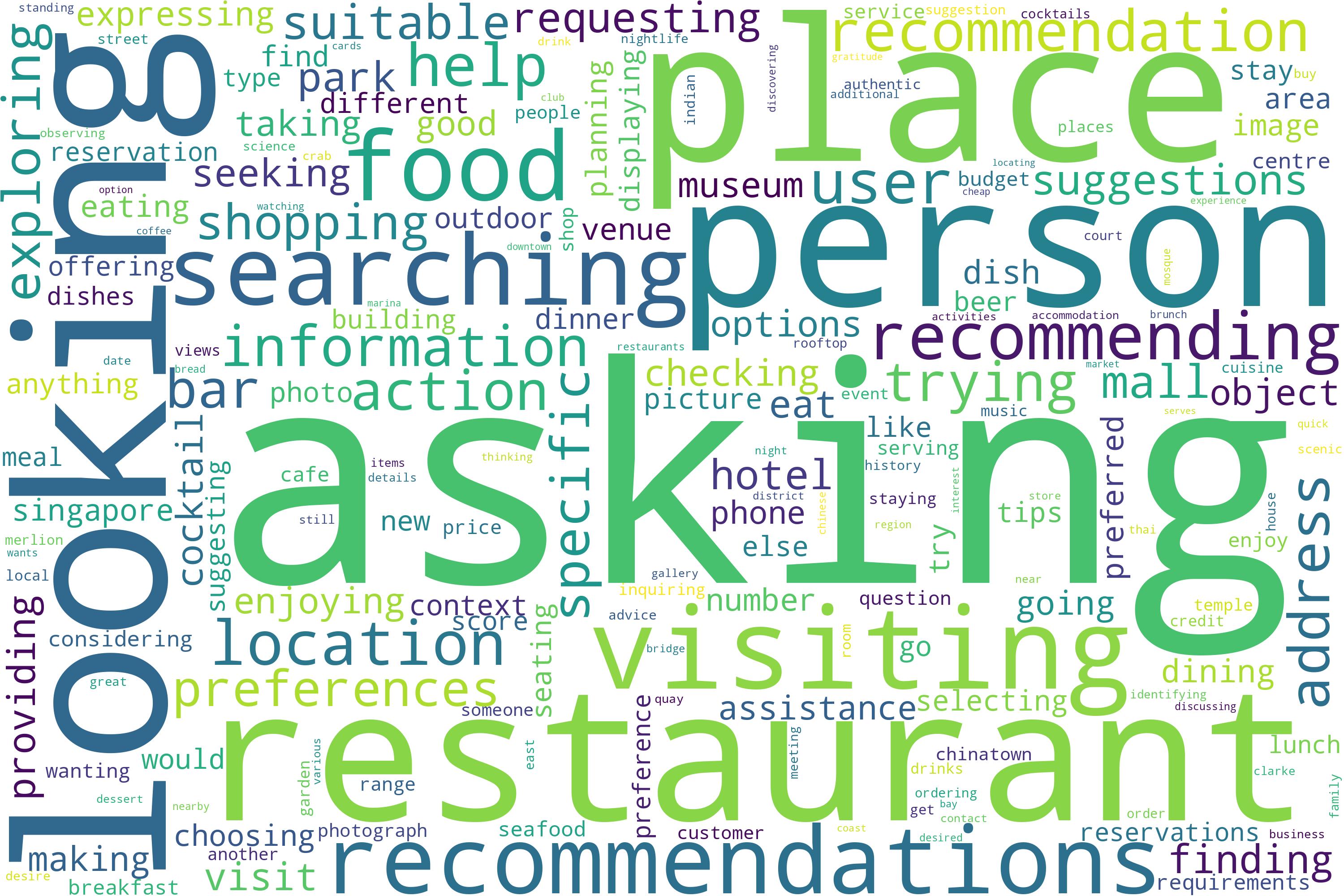}
  }
 \subfigure[Word cloud of keywords.]{
  \includegraphics[scale=0.065]{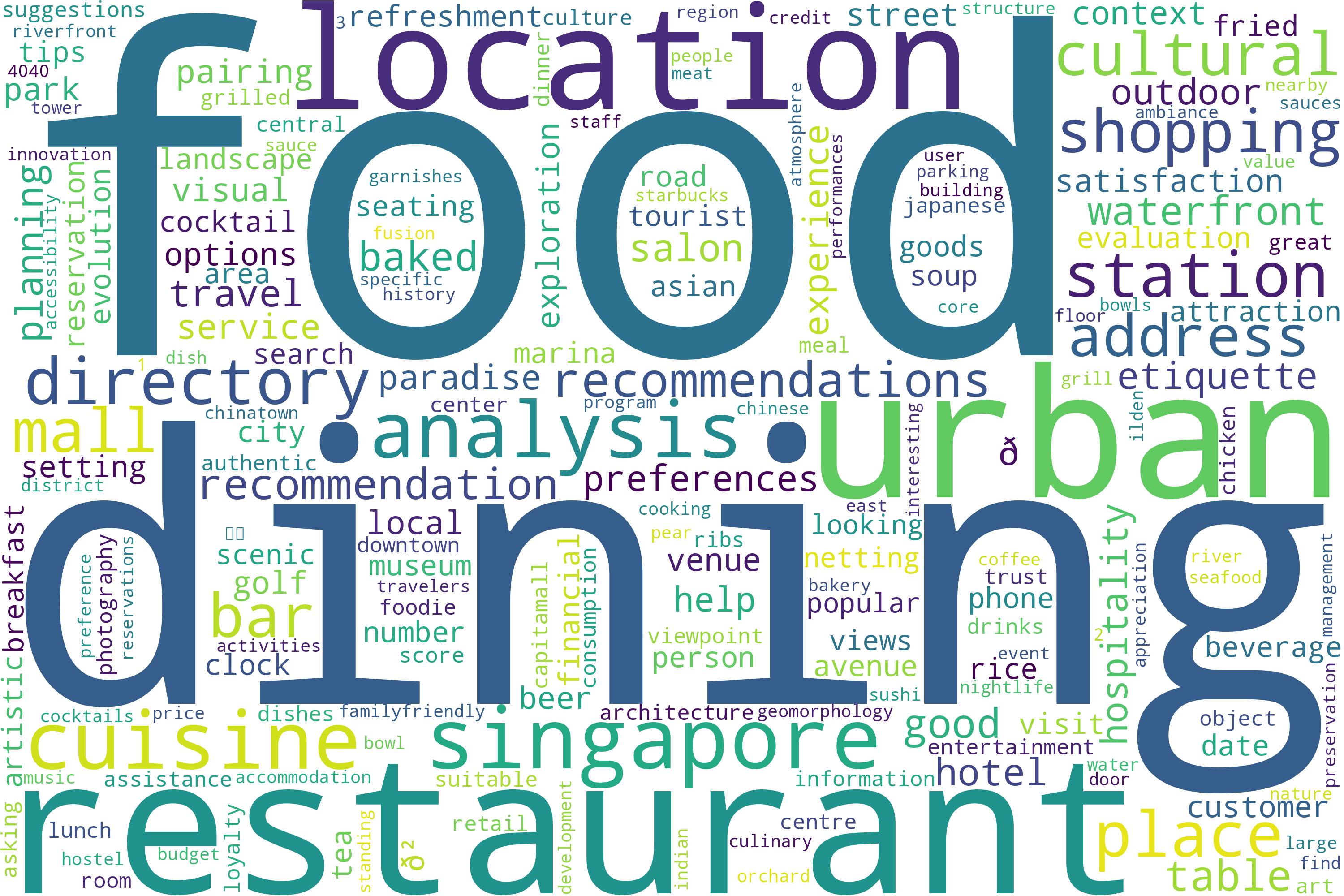}
  }
    \vspace{-1em}
   \caption{Word clouds of learned key clues derived from testing
    samples.}
    \label{chap5:fig_wordcloud}
    % \vspace{-1.5em}
\end{figure}

To  gain deep insights into the learned intention-oriented key clues in multimodal task-oriented dialog systems, we exhibited word clouds of learned user needs and keywords  in Figure~\ref{chap5:fig_wordcloud}. 
The visualization results clearly demonstrate several notable patterns in the linguistic characteristics of user expressions.
As we can see, the  user needs predominantly  cluster around action-driven verbs such as ``asking'' and  ``looking'', and tends  to involve aspects like ``restaurant'', ``place'', and ``person''.
This linguistic distribution strongly aligns with the core communicative purpose of task-oriented dialogues, which fundamentally serve as information-seeking interactions where users systematically combine intention indicators with specific service targets.
Additionally, the most salient keywords prominently feature terms like ``food'', ``dining'',  and ``restaurant''. 
This lexical pattern likely stems from the inherent domain composition of the MMConv  dataset, where food-related dialogues constitute the majority of instances. 
The observed alignment between the extracted keywords and the dataset's domain distribution provides compelling evidence that our DK2R framework effectively identifies and captures domain-specific lexical regularities.
The above observations validate the effectiveness of our  \mbox{intention-oriented} key clues reasoning component.

\begin{table}[!t]
    \centering
    \small
    \caption{Performance comparison between DK2R and its variants with alternative MLLM backbones.}
    \setlength{\tabcolsep}{7mm}{\begin{tabular}{l|rrrrr}
    \hline
        Methods & BLEU-1 & BLEU-2 & BLEU-3 & BLEU-4 & Nist \\ \hline
        
        Yi-VL-$6$B &  $40.55$ & $34.07$ & $30.21$ & $27.46$ & $3.0728$  \\               
        $\text{DK2R}_{\text{Yi-VL-$6$B}}$ & {$41.54$} & {$35.25$} & {$31.50$} & {$28.79$} & {$3.2985$} \\  \hline
        Qwen$2.5$-VL-$7$B & $42.14$ & $35.30$ & $31.27$ & $28.43$ & $3.5536$ \\   
        $\text{DK2R}_{\text{Qwen$2.5$-VL-$7$B}}$ & {$42.29$} & {$36.55$} & {$33.00$} & {$30.35$} & {$3.5452$} \\  \hline      
        {LLaVA-$1.6$-$7$B} & ${45.67}$ & $38.04$ & ${33.63}$ & ${30.41}$ & ${5.0993}$ \\          
         $\text{DK2R}_{\text{LLaVA-$1.6$-$7$B}}$  & {$46.53$} & {$38.70$} & {$34.18$} & {$30.89$} & {$5.2432$} \\ \hline
          {Llama-$3$-$8$B} & ${47.48}$ & $39.48$ & ${34.85}$ & ${31.47}$ & ${5.3308}$ \\
        % \rowcolor{gray!15} \textbf{DK2R} & $\textbf{49.63}$ & $\textbf{41.71}$ & $\textbf{37.10}$ & $\textbf{33.71}$ & $\textbf{5.6856}$ \\ \hline   
         \textbf{DK2R} & $\textbf{49.63}$ & $\textbf{41.71}$ & $\textbf{37.10}$ & $\textbf{33.71}$ & $\textbf{5.6856}$ \\ \hline          
    \end{tabular}}
    \label{chap5:tab_rq4} 
    % \vspace{-1em}
\end{table}

\subsection{On Generalization Capability}
To explore the generalization capability of DK2R, we designed several variants by replacing the backbone model  ${\mathcal{B}_g}(\cdot)$ in Eqn.~(\ref{eq2_resp}) with different MLLMs. Specifically, we replaced 
${\mathcal{B}_g}(\cdot)$ with (1) Yi-VL-$6$B, (2) Qwen$2.5$-VL-$7$B, and (3) LLaVA-$1.6$-$7$B, which are denoted as  $\text{DK2R}_{\text{Yi-VL-$6$B}}$, $\text{DK2R}_{\text{Qwen$2.5$-VL-$7$B}}$, and $\text{DK2R}_{\text{LLaVA-$1.6$-$7$B}}$, respectively.

Table~\ref{chap5:tab_rq4} exhibits the performance comparison between DK2R and its variants with alternative MLLM backbones, revealing the following  key findings.
1) All DK2R-enhanced variants (\emph{i.e.,} $\text{DK2R}_{\text{Yi-VL-$6$B}}$, $\text{DK2R}_{\text{Qwen$2.5$-VL-$7$B}}$, and $\text{DK2R}_{\text{LLaVA-$1.6$-$7$B}}$)  consistently outperform their original backbones (\emph{i.e.,} Yi-VL-$6$B, Qwen$2.5$-VL-$7$B, and LLaVA-$1.6$-$7$B) across most evaluation metrics, demonstrating DK2R's robust generalization across diverse MLLM backbones.
\mbox{2) DK2R} achieves the best performance among all variants (\emph{i.e.,} $\text{DK2R}_{\text{Yi-VL-$6$B}}$, $\text{DK2R}_{\text{Qwen$2.5$-VL-$7$B}}$, and $\text{DK2R}_{\text{LLaVA-$1.6$-$7$B}}$), suggesting that the LLM Llama-$3$-$8$B contributes more substantially to response generation in multimodal task-oriented dialog systems compared to other MLLMs.
And 3) The relative performance ranking among variants (\emph{i.e.,} DK2R$>$ $\text{DK2R}_{\text{LLaVA-$1.6$-$7$B}}$  $>$  $\text{DK2R}_{\text{Qwen$2.5$-VL-$7$B}}$  $>$ $\text{DK2R}_{\text{Yi-VL-$6$B}}$),  which is consistent with that among their original backbones towards textual response generation of   multimodal task-oriented dialog systems.  This experimental phenomenon underscores the critical role of the backbone selection in multimodal task-oriented dialog systems.

\section{Conclusion}
In this paper, we propose a novel dual knowledge-enhanced \mbox{two-stage} reasoner for textual response generation in multimodal dialog systems (named DK2R), which targets solving two fundamental challenges:  (1) dynamic knowledge type selection and (2) intention-response decoupling. 
Specifically, the proposed DK2R consists of  three pivotal components: \textit{context-related dual knowledge extraction}, \textit{probe-driven knowledge type filtering}, and \textit{two-stage reasoning-enhanced response generation}.
Specifically, the context-related dual knowledge extraction component first extracts pertinent both structured attribute and unstructured review knowledge based on the current  context. Thereafter, the probe-driven knowledge type filtering component assesses the relative utility of each knowledge type, determining  the useful type of knowledge. Finally, the two-stage reasoning-enhanced response generation component first summarizes intention-oriented key clues, and then conducts the textual response generation.
Extensive experiments are conducted on a public dataset  to validate  the effectiveness of DK2R. The experimental results demonstrate the superiority of our DK2R over existing methods. 
Besides, ablation studies illustrate that  knowledge type filtering helps retain useful knowledge types and thereby improves textual response generation.
Interestingly, we observe that structured attribute  knowledge contributes more to textual response generation than unstructured review knowledge.
Moreover, we learn that the \mbox{intention-oriented} key clues  reasoning  helps capture user intentions and enhances response generation. 
Thereinto, as for the summarized intention-oriented key clues, user needs exert greater influence than  keywords towards textual  response generation.

Presently, we focus on distinguishing the useful knowledge type (\emph{i.e.,} structured vs. unstructured), but have not yet achieved item-level discrimination of beneficial knowledge elements. In  future work,  we plan to   develop \mbox{fine-grained} knowledge filtering mechanisms to enable more precise knowledge extraction for textual response generation.

\begin{acks}
This work is supported by the National Natural Science Foundation of China (No.: 62376137); the Shandong Provincial Natural Science Foundation (No.:ZR2022YQ59); the National Natural Science Foundation of China (No.: 624B2047). 

% the Shandong Provincial Natural Science Foundation (No.:ZR2022YQ59); National Natural Science Foundation of China (No.: 62236003); the Shenzhen College Stability Support Plan (No.:  GXWD20220817144428005); the National Natural Science Foundation of China (No.: 62376137); and the Shandong Provincial Natural Science Foundation (No.:ZR2022YQ59). 
% Xiaolin Chen acknowledges the support from the China Scholarship Council for pursuing her joint Ph.D. studies at National University of Singapore.

\end{acks}

% \clearpage
% \small
\bibliographystyle{ACM-Reference-Format}
\bibliography{sample-base}

\end{document}